\documentclass[11pt]{article}

\usepackage{amssymb,amsmath,amsthm,bbm}
\usepackage{verbatim,float,url,dsfont}
\usepackage{graphicx,subcaption,psfrag}
\usepackage{algorithm}
\usepackage{algpseudocode}
\usepackage{mathtools,enumitem}
\usepackage{multirow}
\usepackage{ragged2e}
\usepackage{xr-hyper}
\usepackage{caption}
\usepackage{array}

\usepackage[utf8]{inputenc} 
\usepackage[T1]{fontenc}    
\usepackage{booktabs}       
\usepackage{nicefrac}         
\usepackage{microtype}      

\usepackage[dvipsnames]{xcolor} 
\definecolor{hunyuanblue}{HTML}{1E4A8F}
\usepackage{pifont} 
\usepackage{etoc} 
\usepackage{tikz} 
\usepackage{pgfplots}  
\pgfplotsset{compat=1.16}  

\ifdefined\TimesFont 
\usepackage{times} 
\fi

\ifdefined\ParSkip 
\usepackage{parskip} 
\fi

\newtheorem*{assumption*}{\assumptionnumber}
\providecommand{\assumptionnumber}{}


\makeatletter
\newcommand*\rel@kern[1]{\kern#1\dimexpr\macc@kerna}
\newcommand*\widebar[1]{%
  \begingroup
  \def\mathaccent##1##2{%
    \rel@kern{0.8}%
    \overline{\rel@kern{-0.8}\macc@nucleus\rel@kern{0.2}}%
    \rel@kern{-0.2}%
  }%
  \macc@depth\@ne
  \let\math@bgroup\@empty \let\math@egroup\macc@set@skewchar
  \mathsurround\z@ \frozen@everymath{\mathgroup\macc@group\relax}%
  \macc@set@skewchar\relax
  \let\mathaccentV\macc@nested@a
  \macc@nested@a\relax111{#1}%
  \endgroup
}
\makeatother

\usepackage{amsmath,amsfonts,bm,amsthm}

\def\1{\bm{1}}

\def\vc{{\bm{c}}}

\def\vv{{\bm{v}}}

\DeclareMathAlphabet{\mathsfit}{\encodingdefault}{\sfdefault}{m}{sl}
\SetMathAlphabet{\mathsfit}{bold}{\encodingdefault}{\sfdefault}{bx}{n}

\usepackage{hunyuan}

\usepackage[capitalize,noabbrev]{cleveref}
\usepackage{tabularx}
\usepackage[most]{tcolorbox}
\usepackage{wrapfig}
\usepackage{placeins}   
\usepackage{xspace}
\usepackage{fancyhdr}
\usepackage{xurl}
\usepackage{colortbl}   
\usepackage{titletoc}   

\makeatletter
\long\def\@makecaption#1#2{%
  \vskip 10pt
  \setbox\@tempboxa\hbox{#1: #2}%
  \ifdim \wd\@tempboxa >\hsize
    \noindent #1: #2\par   
  \else
    \hbox to\hsize{\hfil\box\@tempboxa\hfil}
  \fi}
\makeatother

\hypersetup{
  colorlinks=true,
  linkcolor=hunyuanblue,
  citecolor=hunyuanblue,
  urlcolor=hunyuanblue,
}

\makeatletter
\def\section{\@startsiction{section}{1}{\z@}{-0.24in}{0.10in}
             {\large\bf\raggedright\color{hunyuanblue}}}
\def\subsection{\@startsection{subsection}{2}{\z@}{-0.20in}{0.08in}
                {\normalsize\bf\raggedright\color{hunyuanblue}}}
\makeatother

\fancypagestyle{plain}{%
  \fancyhf{}%
  \fancyfoot[C]{\thepage}%
}

\fancypagestyle{firststyle}{%
  \fancyhf{}%
  \fancyhead[L]{\raisebox{-18.5mm}[0pt][0pt]{\includegraphics[height=12mm]{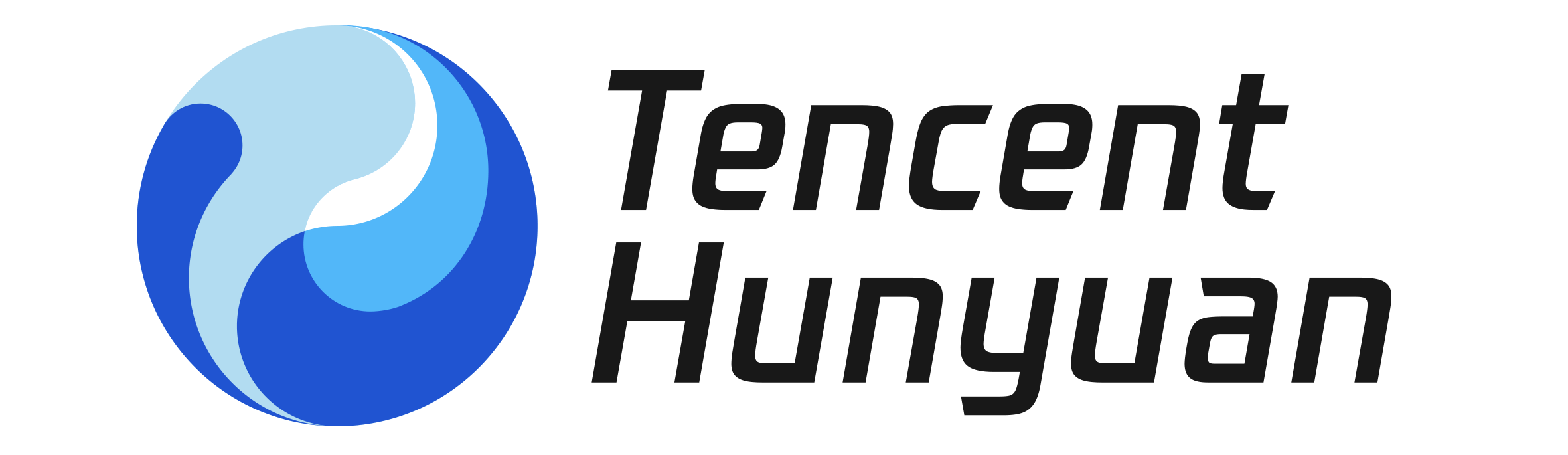}}}%
  \fancyfoot[C]{\thepage}%
}

\makeatletter
\newcommand{\appendixtitle}{%
  \clearpage
  \thispagestyle{plain}%
  \vbox{%
    \hsize\textwidth
    \linewidth\hsize
    \vskip 0.1in
    {\color{hunyuanblue}\hrule height 0.6pt}%
    \vskip 4mm
    \centering
    {\LARGE\bfseries Appendix of \@title\par}%
    \vskip 4mm
    {\color{hunyuanblue}\hrule height 0.6pt}%
    \vskip 0.3in\@minus0.1in
  }%
}
\makeatother

\newcommand{\answerTODO}[1][]{\textcolor{red}{\bf [TODO]}}

\newcolumntype{Y}{>{\raggedright\arraybackslash}X}

\definecolor{caseblue}{RGB}{42, 91, 160}
\definecolor{casebg}{RGB}{246, 248, 252}
\definecolor{caseborder}{RGB}{180, 195, 220}
\definecolor{paperblue}{HTML}{1F77B4}
\definecolor{paperred}{HTML}{D62728}
\definecolor{deepred}{HTML}{B22222}
\definecolor{softred}{HTML}{C44E52}

\newtcolorbox[auto counter, number within=section]{compactcase}[2][]{
  breakable,
  enhanced,
  colback=gray!2,
  colframe=gray!30,
  colbacktitle=gray!12,
  coltitle=black,
  fonttitle=\bfseries,
  title={#2},
  boxrule=0.45pt,
  arc=1mm,
  left=1.5mm,
  right=1.5mm,
  top=1mm,
  bottom=1mm,
  toptitle=0.7mm,
  bottomtitle=0.7mm,
  before skip=0.8em,
  after skip=0.8em,
  label={#1}
}

\definecolor{abstractbg}{HTML}{F0F7FC}

\title{TempAct: Advancing Temporal Plausibility in Autoregressive Video Generation via Planner-Executor Reinforcement Learning}

\begin{document}

\thispagestyle{firststyle}
\vspace*{0.25cm}
{\color{hunyuanblue}\hrule height 0.6pt}
\vskip 6mm
\begin{center}
{\LARGE\bfseries TempAct: Advancing Temporal Plausibility in Autoregressive Video Generation via 
Planner-Executor RL\par}
\end{center}
\vskip 3mm
{\color{hunyuanblue}\hrule height 0.6pt}
\vskip 6mm
\begin{center}
\textbf{Jing Wang}$^{1,2,*}$ \quad
\textbf{Xiangxin Zhou}$^{2,*}$ \quad
\textbf{Jiajun Liang}$^{3,\mathparagraph}$  \quad
\textbf{Kaiqi Liu}$^{4}$\\
\textbf{Wanyuan Pang}$^{5}$ \quad
\textbf{Zhenyu Xie}$^{1}$ \quad
\textbf{Tianyu Pang}$^{2,\ddagger}$ \quad
\textbf{Xiaodan Liang}$^{1,\ddagger}$
\\[8pt]
$^1$Shenzhen Campus of Sun Yat-Sen University, $^2$Tencent Hunyuan, \\ $^3$Tsinghua University, $^4$Peking University, $^5$USTB \\
{\small $^*$Equal contribution \quad
$^\mathparagraph$Project Lead \quad
$^\ddagger$Corresponding author}
\end{center}
\vskip 6mm

\begin{tcolorbox}[
  colframe=abstractbg,
  colback=abstractbg,
  boxrule=0pt,
  arc=2mm,
  enhanced,
  top=12pt,
  bottom=12pt,
  left=15pt,
  right=15pt,
  width=\textwidth,
]
\textbf{Abstract.}\quad
Autoregressive (AR) video diffusion models enable low-latency streaming generation by synthesizing videos chunk by chunk with cached visual context, but this chunk-wise formulation makes temporal instruction following ambiguous. A single global prompt does not specify which sub-event should be realized in each chunk, while naively switching to step-wise prompts often leads to delayed reactions, blended step semantics, and error propagation across prompt transitions. These failures are difficult to address with supervised fine-tuning or distillation alone: SFT suffers from exposure bias, while rollout-based distillation still optimizes low-level denoising or teacher-distribution matching rather than directly enforcing action ordering and prompt-transition correctness. We propose \textbf{TempAct}, \textbf{a planner--executor reinforcement learning} framework that jointly optimizes temporal decomposition and step-conditioned execution for temporally plausible AR video generation. TempAct uses an LLM planner to explore span-aware step prompts that are executable by the video model, and trains an AR diffusion executor to follow these prompts under its own generated histories. Its key mechanism is hierarchical group exploration: candidate plans form planning groups, and each plan induces an execution group of multiple continuations from a shared visual context, enabling plan-level credit assignment for long-horizon temporal outcomes and executor-level credit assignment for prompt-switch behavior. We further design hierarchical rewards that combine plan-quality and full-video temporal feedback for the planner with local transition-level step-following rewards, aesthetic regularization, and KL constraints for the executor. Experiments on Self-Forcing and LongLive show that TempAct improves temporal consistency while preserving overall visual quality.

\vskip 8pt
\textbf{Date:} June 26, 2026 \\
\textbf{Code:} \url{https://github.com/jingw193/TempAct} \\
\textbf{Project Page:} \url{https://jingw193.github.io/TempAct/}
\end{tcolorbox}

\section{Introduction}

Video diffusion models~\citep{kling2024,wan2025,chen2025skyreels,jin2025pyramidal,wang2025pt,Seedance2026, wang2026wisa, liu2026mavin} have advanced rapidly in visual fidelity and motion realism, but deploying them in interactive applications requires more than high-quality offline generation: the model must generate streams with low latency while maintaining temporal coherence over long horizons. A common response to this demand is to distill pretrained video diffusion models into few-step autoregressive (AR) video generators~\citep{huang2026self, cui2025self, yang2025longlive, zhu2026causal}, often using distribution-matching distillation (DMD)~\citep{yin2024one} to preserve generation quality while reducing sampling cost. In this formulation, the model no longer synthesizes an entire video at once; instead, it generates one chunk after another, using previously generated frames as causal visual context. This context is stored through mechanisms such as attention states, KV caches, or visual caches, allowing subsequent chunks to be produced efficiently while preserving continuity with the past.

\begin{figure}[t]
    \centering
    \includegraphics[width=1.0\linewidth]{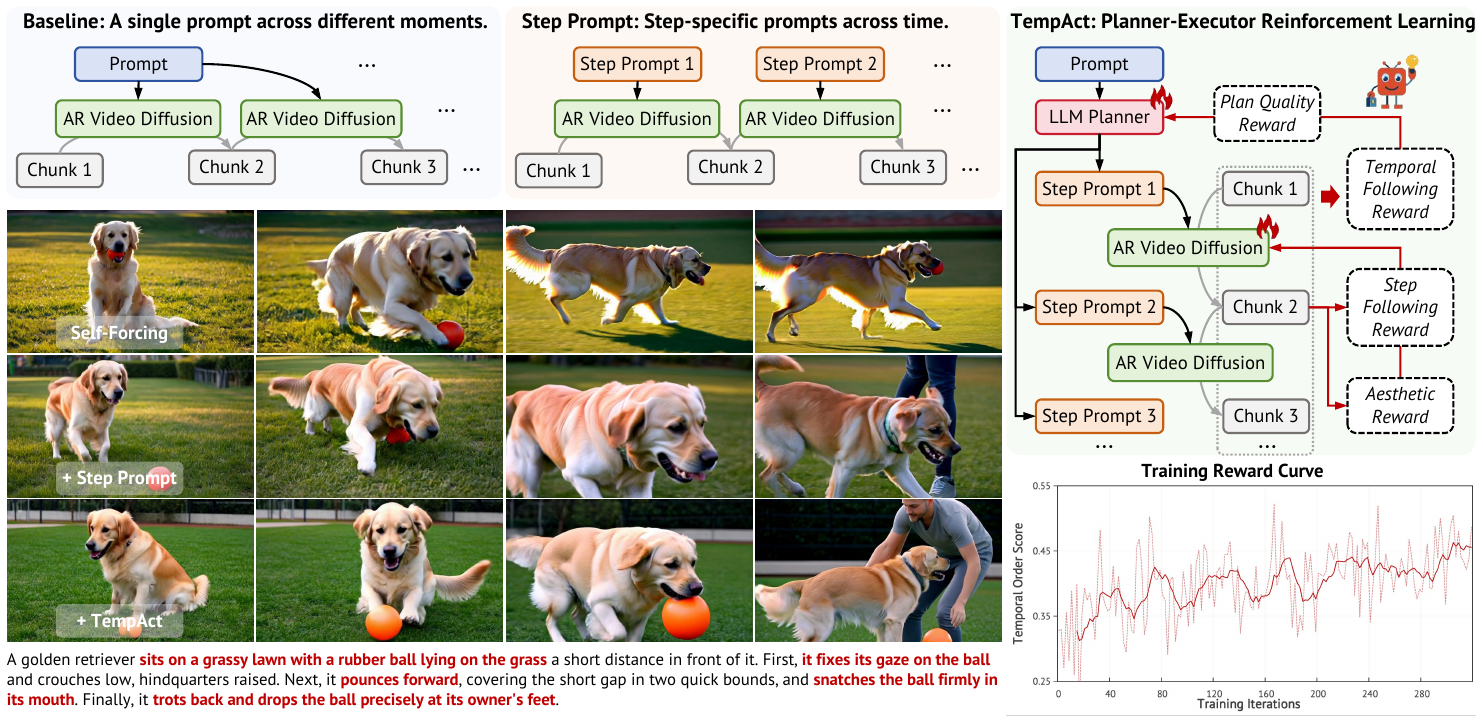}
    \vspace{-9mm}
    \caption{\textbf{Overview and Motivation of TempAct.} \textbf{Framework.} Single-prompt AR generation conditions every chunk on the same global instruction, while step-prompt generation provides explicit stage-wise conditions but still relies on a fixed executor. TempAct introduces a planner--executor RL framework that jointly optimizes temporal decomposition and prompt-transition execution. \textbf{Qualitative comparison}. Compared with single-prompt and step-prompt baselines, TempAct produces more faithful event progression under temporally complex instructions. \textbf{Training dynamics}. The increasing reward curve shows that hierarchical planner--executor optimization provides effective learning signals for temporal plausibility.}
    \label{fig:intro_teaser}
\end{figure}

This chunk-wise formulation improves efficiency, but it can also make temporal instruction following ambiguous. In most AR generators, each chunk is typically conditioned on the same global prompt. As a result, the model knows the full instruction but not which part of the instruction should be realized in the current chunk. For example, in the dog example in Figure~\ref{fig:intro_teaser}, the instruction requires the dog to first sit and then pick up the ball and run. However, the single-prompt Self-Forcing baseline prematurely depicts the dog holding the ball while sitting at the beginning, causing an event from a later stage to appear too early. We refer to this failure as \textit{temporal confusion}: the video may look reasonable at each frame, but \textbf{the event order is collapsed or blurred across chunks}. A straightforward fix is to replace the global prompt with step-wise prompts that specify the active subgoal for each stage. However, this shifts the difficulty to handling prompt transitions. As illustrated in Figures~\ref{fig:intro_teaser}, \ref{fig:self_forcing_compare}, and~\ref{fig:longlive_compared}, when the prompt changes, the AR executor may keep following the previous action, blend adjacent step semantics, or carry errors from earlier chunks into later ones.

These failures are difficult to eliminate with supervised fine-tuning (SFT) or distillation alone. SFT suffers from exposure bias because it trains on clean teacher-forced histories but must infer from imperfect rollouts at test time~\citep{schmidt2019generalization,ning2024elucidating}. Rollout-based distillation methods such as DMD can reduce this mismatch, but both SFT and distillation still optimize low-level denoising or teacher-distribution matching~\citep{lipman2022flow}, rather than explicitly enforcing action ordering or prompt-transition correctness. This motivates reinforcement learning~\citep{shao2024deepseekmath, liu2025flowgrpo}: by exploring generated plans and videos and scoring them with LLM/VLM feedback~\citep{guo2025deepseek,jaech2024openai,Qwen3-VL}, RL can directly reward \textbf{executable temporal plans}, \textbf{successful prompt-switch execution}, and \textbf{long-horizon consistency}.

Motivated by this, we propose \textbf{TempAct}, a planner--executor reinforcement learning framework for temporally plausible AR video generation. Our key observation is that temporally complex generation requires solving two coupled problems: first determining \emph{what} should happen at each stage, and then deciding \emph{how} to realize that stage under autoregressive visual history. Executor-only training mainly improves the second problem, i.e., how well the video model follows a given condition, but it cannot revise an ambiguous or poorly timed temporal decomposition. A planner--executor framework therefore provides a more general solution: the planner can organize event order, step granularity, and state changes, while the executor learns to realize prompt transitions without drifting from the generated visual context. TempAct instantiates this idea by using an LLM planner to explore span-aware decompositions of a global instruction and an autoregressive video diffusion executor to roll out chunks conditioned on the active step and accumulated visual history. Crucially, \textbf{the LLM is not used as a fixed preprocessing module}; instead, TempAct optimizes both the planner and executor from generated trajectories, allowing the planner to learn which temporal plans are executable by the video model and the executor to learn how to realize prompt switches under its own imperfect histories.

To make this joint optimization effective, TempAct introduces a \textbf{hierarchical group exploration strategy} in which execution groups are nested under planning groups. For each global prompt, the planner samples a group of candidate temporal plans, covering different action orders, temporal granularities, and state descriptions. Conditioned on each candidate plan, the executor constructs a plan-specific execution group: it first generates a shared context from an earlier step, and then samples multiple continuations after the step transition under that same context. This nested structure provides aligned credit assignment at two levels: comparisons across planning groups reward decompositions that lead to better full-video temporal outcomes, while comparisons within each execution group isolate prompt-switch execution from variations in preceding visual history. Rewards follow the same hierarchy, combining plan-quality and full-video temporal-following feedback for the planner with local transition-level rewards and aesthetic regularization for the executor.

Empirically, we evaluate temporal plausibility using a Temporal-Following Score that measures whether generated videos follow the intended event sequence. TempAct improves this score across Self-Forcing and LongLive backbones on temporally complex video generation tasks, with consistent gains under both in-domain and out-of-domain VLM evaluation, as well as human evaluation.
\section{Related Works}

\subsection{Autoregressive Video Generation}

Autoregressive (AR) video generation enables streaming synthesis by producing frames or chunks sequentially, but teacher-forced AR diffusion models suffer from train-test mismatch and error accumulation when conditioned on their own histories~\citep{hu2024acdit,gao2024ca2}. Recent work improves AR generation through causal distillation, rollout-based training, and memory management: CausVid converts bidirectional diffusion transformers into causal generators with video DMD and KV caching~\citep{yin2024slow,yin2024one}; Self-Forcing trains with autoregressive rollouts and video-level feedback to mitigate exposure bias~\citep{huang2026self,cui2025self}; subsequent methods manage contextual memory for more consistent long rollouts~\citep{ji2025memflow,ji2026forcing,yi2025deep,yang2026anchor}; LongLive further introduces KV-recache, streaming long tuning, and frame-sink attention for interactive long-video generation~\citep{yang2025longlive}; and Causal Forcing bridges the architectural gap between bidirectional teachers and causal students by using an AR teacher for ODE initialization before DMD post-training~\citep{zhu2026causal,zhao2026causal}. These methods improve streaming efficiency and rollout stability under a given text condition, whereas TempAct focuses on instruction-guided temporal plausibility by jointly optimizing temporal planning and prompt-transition execution with hierarchical reinforcement learning.

\subsection{Reinforcement Learning for Generative Models}

The success of group-relative policy optimization in language models has inspired growing interest in reinforcement learning for generative models. For diffusion and flow-matching models, Flow-GRPO and DanceGRPO introduce stochastic sampling paths that make GRPO-style policy updates applicable to visual generation~\citep{liu2025flowgrpo,xue2025dancegrpo}. Subsequent studies improve the efficiency and stability of online diffusion RL through better sampling schemes, credit assignment, clipping strategies, and regularization~\citep{li2025mixgrpo,wang2025coefficients,li2025branchgrpo,li2026aegpo,wang2026grpo,he2025gardo, wang2026tagrpo}. Beyond these optimizer and sampler designs, video feedback learning shows that preference and semantic rewards can improve video generation quality and alignment~\citep{liu2026improving}.

Recent work further applies RL to autoregressive video generation. AR-CoPO~\citep{he2026ar} uses chunk-level forking and localized contrastive updates. Astrolabe~\citep{zhang2026astrolabe} aligns distilled AR models with forward-process negative-aware fine-tuning and streaming local-window updates. KVPO~\citep{zhang2026kvpo} introduces ODE-native GRPO with KV-cache semantic exploration and velocity-space contrastive objectives. Together, these methods advance executor-level preference alignment for streaming AR generators, but they do not explicitly model how complex instructions should be temporally organized, leaving event ordering and long-horizon semantic dependencies under-specified. In contrast, TempAct performs planner--executor RL over temporally complex instructions, using an LLM planner to explore executable temporal decompositions and an AR executor to learn prompt-switch execution under hierarchical credit assignment, offering a new perspective on long-horizon temporal plausibility.

\section{Preliminaries}

\noindent\textbf{Autoregressive Video Diffusion Models.}
Autoregressive (AR) video diffusion models provide the basic streaming formulation used by recent long-video generation methods such as Self-Forcing \citep{huang2026self} and LongLive \citep{yang2025longlive}. Given a text condition $\vc$, an AR video model represents a video sequence $\mathbf{x}_{1:K}=\{\mathbf{x}_1,\ldots,\mathbf{x}_K\}$ through the causal factorization
\begin{equation}
    p_\theta(\mathbf{x}_{1:K}\mid \vc)
    = \prod_{i=1}^{K} p_\theta(\mathbf{x}_i \mid \mathbf{x}_{<i}, \vc),
\end{equation}
where $\mathbf{x}_i$ denotes the $i$-th generated frame or video chunk, and $\mathbf{x}_{<i}=\{\mathbf{x}_1,\ldots,\mathbf{x}_{i-1}\}$ denotes the previously generated frames or chunks used as causal visual context. Generation then proceeds sequentially, sampling each $\mathbf{x}_i$ conditioned on $\mathbf{x}_{<i}$ and $\vc$. 

Each conditional distribution $p_\theta(\mathbf{x}_i \mid \mathbf{x}_{<i}, \vc)$ is modeled by flow matching \citep{lipman2022flow,liu2022flow}. 
To train the model, for the $i$-th frame or chunk $\mathbf{x}_i$, a noise endpoint $\boldsymbol{\epsilon}_i\sim\mathcal{N}(\mathbf{0},\mathbf{I})$ and a timestep $t\sim\mathcal{U}(0,1)$ are sampled, and the intermediate state is defined as
$\mathbf{x}_i^t = (1-t)\mathbf{x}_i + t\boldsymbol{\epsilon}_i$.
The model predicts the velocity field $\vv_\theta(\mathbf{x}_i^t,t\mid \mathbf{x}_{<i},\vc)$ from the noisy state, timestep, text condition, and causal visual context. The final training loss is the conditional flow-matching (CFM) objective, defined as 
\begin{equation}
    \mathcal{L}_{\mathrm{CFM}}(\theta)
    = \mathbb{E}_{i,t,\mathbf{x}_i,\boldsymbol{\epsilon}_i}
    \left[\left\|\vv_\theta(\mathbf{x}_i^t,t\mid \mathbf{x}_{<i},\vc)
    - (\boldsymbol{\epsilon}_i-\mathbf{x}_i)\right\|_2^2\right].
\end{equation}

The main difference among AR training schemes lies in how the historical context is constructed. Teacher forcing (TF)~\citep{williams1989learning,rasul2021autoregressive} conditions each prediction on clean ground-truth history $\mathbf{x}_{<i}$. Diffusion forcing (DF)~\citep{chen2024diffusion} improves robustness by corrupting the historical frames with independently sampled noise levels, so the model learns to denoise under noisy contexts $\{\mathbf{x}_j^{t_j}\}_{j<i}$. However, both TF and DF still build the history from data rather than from the model's own generated frames, causing exposure bias when the model must condition on its own outputs at inference time. Self-Forcing~\citep{huang2026self} addresses this mismatch by training on rollouts from the current model, $\mathbf{x}_{1:K}^\theta \sim \prod_{i=1}^{K}p_\theta(\mathbf{x}_i\mid \mathbf{x}_{<i},\vc)$, and distilling teacher velocity predictions along these self-generated trajectories.

\noindent\textbf{GRPO and Flow-GRPO.}
Group Relative Policy Optimization (GRPO) is a critic-free policy optimization algorithm for improving LLM reasoning ability \citep{shao2024deepseekmath}. For a prompt $q$, the current policy $\pi_\theta$ samples a group of $G$ responses $\{o_i\}_{i=1}^{G}$, each scored by a reward function $r_i=r(q,o_i)$. GRPO estimates the advantage of each response by normalizing rewards within the same group:
\begin{equation}
    A_i = \frac{r_i - \mathrm{mean}(\{r_j\}_{j=1}^{G})}
    {\mathrm{std}(\{r_j\}_{j=1}^{G}) + \epsilon}.
\end{equation}
The policy is then updated with a PPO-style clipped objective and a reference-policy regularizer:
\begin{equation}
    \mathcal{L}_{\mathrm{GRPO}}(\theta)
    = - \mathbb{E}_{q,\{o_i\}}
    \left[\frac{1}{G}\sum_{i=1}^{G}
    \min\left(\rho_i A_i,\mathrm{clip}(\rho_i,1-\epsilon_c,1+\epsilon_c)A_i\right)
    - \beta D_{\mathrm{KL}}(\pi_\theta \| \pi_{\mathrm{ref}})\right],
\end{equation}
where $\rho_i=\pi_\theta(o_i\mid q)/\pi_{\theta_{\mathrm{old}}}(o_i\mid q)$ and $\pi_{\mathrm{ref}}$ is a fixed reference policy. The key property of GRPO is that relative comparison among samples from the same prompt provides a low-cost baseline, avoiding the extra critic model required by PPO \citep{schulman2017proximal}. Group Sequence Policy Optimization (GSPO) extends this group-relative principle to sequence-level policy updates by normalizing the importance ratio over the generated sequence, providing a stable update rule for long-form LLM generations~\citep{zheng2025group}.

Flow-GRPO extends group-relative policy optimization to diffusion and flow-matching generators~\citep{liu2025flowgrpo}. Since standard flow-matching samplers follow deterministic ODE trajectories, Flow-GRPO uses an ODE-to-SDE conversion to recast generation as a stochastic policy, making it compatible with GRPO-style optimization. In flow-matching RL, however, the importance-ratio distribution can be biased across denoising timesteps, which weakens the intended clipping constraint and may lead to over-optimization. GRPO-Guard~\citep{wang2026grpo} addresses this issue with regulated clipping, including ratio normalization $\hat{\rho}=\mathrm{RatioNorm}(\rho)$ that recenters and stabilizes the timestep-wise ratio distribution before applying the clipped surrogate. 

\section{Method}

\begin{figure}[t]
    \centering
    \includegraphics[width=\linewidth]{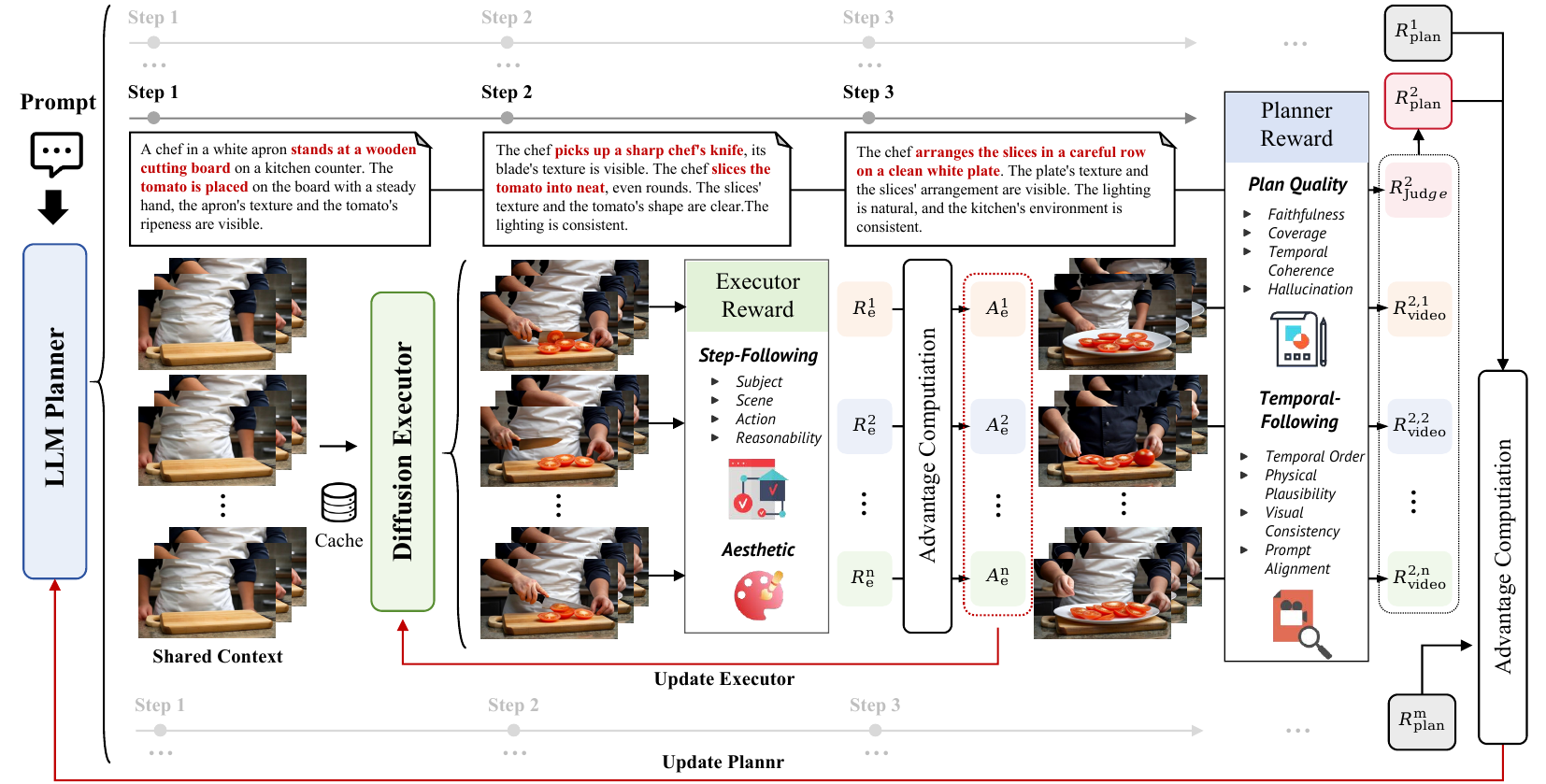}
    \caption{Overview of TempAct. An LLM planner samples span-aware temporal decompositions of a global instruction, while an autoregressive video executor rolls out shared contexts and multiple continuations under the corresponding step prompts. The nested planning--execution groups support hierarchical credit assignment, with plan-level rewards guiding executable decompositions and local transition-level rewards improving prompt-switch execution.}
    \label{fig:method}
\end{figure}

To address temporal confusion and weak step following in autoregressive video generation, we propose TempAct, a planner--executor reinforcement learning framework for instruction-guided temporal plausibility. As shown in Figure~\ref{fig:method}, TempAct uses an LLM planner $\pi_\phi$ to decompose a global prompt into span-aware step prompts and an autoregressive diffusion executor $p_\theta$ to realize these steps under generated visual context. Section~\ref{sec:hierarchical_pipeline} introduces the hierarchical planner--executor pipeline, Section~\ref{sec:joint_rl} presents the joint RL objective, and Section~\ref{sec:reward_design} describes the multi-level reward design.

\subsection{Hierarchical Planner-Executor Pipeline}
\label{sec:hierarchical_pipeline}

AR video diffusion models suffer from temporal confusion when conditioned on a global prompt, while simply switching to step-wise prompts still leaves the executor with limited step-following ability. Moreover, supervised fine-tuning is insufficient to address these failures, as it suffers from exposure bias and optimizes flow-matching objectives that are only indirectly aligned with temporal semantic correctness. We therefore jointly optimize the planner and executor via reinforcement learning: the planner is updated with GSPO from plan-level and video-level advantages, and the executor is updated with Flow-GRPO from local step-following and aesthetic advantages.

\noindent\textbf{LLM-based Step Planning and Prompt Enhancement.}
Given a global instruction $\vc$ and a target latent video length $K$, we assign each step prompt a temporal span of $L$ latent frames.
The planner $\pi_\phi$ therefore produces a temporally ordered plan $\mathbf{s}=\{s_1,\ldots,s_J\}$ with $J=\lceil K/L\rceil$, where each step prompt $s_j$ describes the action to be executed and the expected visual state during its assigned $L$-frame span.
For example, generating $30$ latent frames with $L=6$ requires the LLM to decompose the prompt into $J=5$ step prompts. Rather than using a deterministic decomposition, we sample a group of candidate plans from the LLM policy,
\begin{equation}
    \mathbf{s}^{m} \sim \pi_\phi(\mathbf{s}\mid \vc), \quad m=1,\ldots,M,
\end{equation}
where $M$ is the planning-level group size. Each sampled plan contains both a step decomposition and prompt enhancement: the original temporal instruction is split into span-aware subgoals, and each subgoal is rewritten with explicit objects, actions, and state constraints for its assigned latent-frame interval. This stochastic planning stage allows the LLM to explore alternative temporal decompositions and receive reward feedback on which plans are more executable for the video generator. We further apply step-prompt smoothing for execution: instead of conditioning the diffusion model only on the current step, each span is represented by a smoothed prompt $\tilde{s}_j=\mathrm{Smooth}(s_j,s_{j+1})$ that exposes the executor to both the current subgoal and the upcoming subgoal. This transition-aware prompt reduces abrupt semantic changes between adjacent spans and empirically prevents large performance drops caused by hard prompt switching.

\noindent\textbf{Hierarchical Autoregressive Video Sampling.}
For each sampled plan $\mathbf{s}^{m}$, the executor $p_\theta$ generates $K$ latent video frames autoregressively under the smoothed step prompt assigned to each temporal span, where $K$ is the target latent length defined above. To train the executor at a step transition from $s_j^{m}$ to $s_{j+1}^{m}$, we first generate a shared context $\mathbf{x}_{1:k}^{m}$ up to the end of the current span, where $k=\min(jL,K)$. Each frame $i$ is conditioned on the smoothed prompt of its own span $\sigma(i)=\lceil i / L \rceil$, where $\tilde{s}_{\sigma(i)}^{m}=\mathrm{Smooth}(s_{\sigma(i)}^{m},s_{\sigma(i)+1}^{m})$,
\begin{equation}
    \mathbf{x}_{1:k}^{m} \sim \prod_{i = 1}^{k} p_\theta(\mathbf{x}_i \mid \mathbf{x}_{<i}, \tilde{s}_{\sigma(i)}^{m}),
\end{equation}
then sample $N$ possible continuations $\{\mathbf{y}_{k+1:K}^{m,n}\}_{n=1}^{N}$ that complete the remaining video from span $j+1$ to the end. For each future latent frame, the executor conditions on the shared context, previously generated continuation frames, and the smoothed prompt assigned to that frame's temporal span,
\begin{equation}
    \mathbf{y}_{k+1:K}^{m, n} \sim \prod_{i = k+1}^{K} p_\theta(\mathbf{y}_{i}^{m,n} \mid \mathbf{x}_{1:k}^{m}, \mathbf{y}_{k+1:i-1}^{m,n}, \tilde{s}_{\sigma(i)}^{m}),
    \quad n=1,\ldots,N.
\end{equation}
This yields a hierarchical group $\{\mathbf{s}^{m},\mathbf{x}_{1:k}^{m},\{\mathbf{y}_{k+1:K}^{m,n}\}_{n=1}^{N}\}_{m=1}^{M}$: the outer group compares different span-aware LLM plans, while the inner group compares different full-video continuations starting from the same context. The full continuations are used for reward evaluation, but the executor update is applied only to the first chunk after the prompt switch, so the credit assignment focuses on prompt-switch execution rather than all future spans.

\subsection{Joint RL of Planner and Executor}
\label{sec:joint_rl}

The sampled trajectories are scored by rewards defined in the next subsection. Let $R_{\mathrm{plan}}^{m}$ denote the planner-level reward for plan $\mathbf{s}^{m}$, computed from plan executability and video-level temporal instruction following. Let $R_{\mathrm{exec}}^{m,n}$ denote the executor-level reward computed only on the first chunk after the prompt switch, measuring local step following and visual quality at the prompt-switch moment. We compute group-relative advantages separately at the two levels,
\begin{equation}
    A_{\mathrm{plan}}^{m} = \frac{R_{\mathrm{plan}}^{m}-\mu_{\mathrm{plan}}}{\sigma_{\mathrm{plan}}+\epsilon},
    \quad
    A_{\mathrm{exec}}^{m,n} = \frac{R_{\mathrm{exec}}^{m,n}-\mu_{\mathrm{exec}}^{m}}{\sigma_{\mathrm{exec}}^{m}+\epsilon},
\end{equation}
where $\mu_{\mathrm{plan}}$ and $\sigma_{\mathrm{plan}}$ are computed over the $M$ sampled plans, while $\mu_{\mathrm{exec}}^{m}$ and $\sigma_{\mathrm{exec}}^{m}$ are computed over the $N$ transition segments sharing the same plan and context. The LLM planner is updated with Group Sequence Policy Optimization (GSPO)~\citep{zheng2025group}. Following GSPO, we define the sequence importance ratio for a sampled plan as
\begin{equation}
    \rho_{\phi}^{m}=\left(\frac{\pi_\phi(\mathbf{s}^{m}\mid \vc)}{\pi_{\phi_{\mathrm{old}}}(\mathbf{s}^{m}\mid \vc)}\right)^{1/|\mathbf{s}^{m}|},
\end{equation}
where $|\mathbf{s}^{m}|$ denotes the number of generated planner tokens. The planner objective is
\begin{equation}
    \mathcal{L}_{\mathrm{planner}}(\phi)
    = -\mathbb{E}\left[\frac{1}{M}\sum_{m=1}^{M}
    \min\left(\rho_{\phi}^{m}A_{\mathrm{plan}}^{m},
    \mathrm{clip}(\rho_{\phi}^{m},1-\epsilon_{\mathrm{p}},1+\epsilon_{\mathrm{p}})A_{\mathrm{plan}}^{m}\right)\right],
\end{equation}
 We regularize the planner with $\beta_{\mathrm{p}}D_{\mathrm{KL}}(\pi_\phi\|\pi_{\mathrm{ref}})$. Unlike token-level PPO ratios, this sequence-level ratio assigns one relative update weight to the entire LLM plan, matching the fact that rewards are obtained after executing the complete temporal decomposition.

The diffusion executor is optimized with a clipped Flow-GRPO objective on the first chunk after the transition to span $j+1$, $\mathbf{y}_{k+1:k+t}^{m,n}$, where $t$ is the chunk size, rather than on the entire remaining continuation. This choice is important because group-relative policy gradients compare samples under the same conditioning state: within this transition chunk, all candidates share the same preceding context and newly activated step prompt, whereas later chunks are generated from sample-specific histories and therefore correspond to different policy inputs. It also improves credit assignment by applying the RL update exactly at the prompt-switch moment, where the action for the new subgoal is decided and any execution error can propagate autoregressively to subsequent chunks. The executor objective is:
\begin{equation}
    \mathcal{L}_{\mathrm{exec}}(\theta)
    = -\mathbb{E}_{m,\{\mathbf{y}_{k+1:k+t}^{m,n}\}_{n=1}^{N}}
    \left[\frac{1}{N}\sum_{n=1}^{N}
    \min\left(
    \hat{\rho}_{\theta}^{m,n}A_{\mathrm{exec}}^{m,n},
    \mathrm{clip}(\hat{\rho}_{\theta}^{m,n},1-\epsilon_{\mathrm{e}},1+\epsilon_{\mathrm{e}})
    A_{\mathrm{exec}}^{m,n}
    \right)\right].
\end{equation}
Here $\rho_{\theta}^{m,n}$ is the likelihood ratio of the denoising trajectory for $\mathbf{y}_{k+1:k+t}^{m,n}$, and $\hat{\rho}_{\theta}^{m,n}=\mathrm{RatioNorm}(\rho_{\theta}^{m,n})$ is the normalized ratio defined above. In practice, we also add the KL regularizer $\beta_{\mathrm{e}}D_{\mathrm{KL}}(p_\theta(\mathbf{y}_{k+1:k+t} \mid\mathbf{x}_{1:k},\tilde{s}_{j+1})\|p_{\mathrm{ref}}(\mathbf{y}_{k+1:k+t} \mid\mathbf{x}_{1:k},\tilde{s}_{j+1}))$ to constrain the transition-span policy from drifting too far from the reference diffusion policy. Together, these updates encourage the planner to discover step prompts that are temporally coherent and executable, and the executor to realize the active subgoal at the prompt-switch moment under imperfect autoregressive context.

\subsection{Multi-level Reward Design}
\label{sec:reward_design}

\noindent\textbf{Planner Rewards.}
The planner is rewarded from two complementary perspectives: whether the generated decomposition is a high-quality temporal plan before execution, and whether executing this plan leads to videos that follow the original temporal instruction.

\begin{itemize}
    \item \textit{Plan Quality Score.} We use Qwen3-8B~\citep{qwen3} as an LLM-based plan judge to evaluate each sampled plan $\mathbf{s}^{m}$ against the original prompt $\vc$. The judge focuses on faithfulness to the original instruction, coverage of all requested events, temporal coherence across steps, and hallucination avoidance. This produces a plan-quality reward $R_{\mathrm{judge}}^{m}$ that encourages the planner to preserve the original event structure, distribute actions across meaningful visual stages, and avoid unsupported additions.
    \item \textit{Temporal-Following Score.} Conventional semantic alignment metrics, such as CLIP score, ImageReward, and VideoAlign~\citep{liu2026improving}, are often insensitive to event ordering and long-range temporal dependencies. We therefore use a VLM-based evaluator to score each executed video. Given sampled video frames and the original prompt, the evaluator not only predicts a temporal-order score, but also assesses physical consistency, visual consistency, and text-video consistency at both frame-level and video-level granularity. These auxiliary criteria prevent the policy from optimizing temporal order alone while ignoring visual plausibility or prompt faithfulness. For each sampled plan $\mathbf{s}^{m}$, we average the video-level scores over its $N$ execution continuations to obtain $R_{\mathrm{video}}^{m}=\frac{1}{N}\sum_{n=1}^{N}R_{\mathrm{video}}^{m,n}$. Due to the high cost of large-scale training-time evaluation with stronger proprietary VLMs (like Gemini), we use Qwen3-VL-8B as the training-time video evaluator.
\end{itemize}

The final planner-level reward combines the textual plan-quality signal and the realized video-level temporal-following signal as $R_{\mathrm{plan}}^{m}$, balancing decomposition quality with executable temporal consistency. Detailed instructions, scoring rules, aggregation weights, and scoring examples are provided in Appendix~\ref{app:reward_details}.

\noindent\textbf{Executor Rewards.}
The executor is rewarded from two complementary perspectives: whether it correctly realizes the newly activated step prompt after a prompt switch, and whether this semantic optimization preserves visual quality. Both signals are computed only on the first transition span $\mathbf{y}_{k+1:k+L}^{m,n}$ so that the reward remains aligned with the transition-span Flow-GRPO update.

\begin{itemize}
    \item \textit{Local Step-Following Score.} Similar to the temporal-following score, we use a VLM-based evaluator to assess step-following performance. For each first transition span $\mathbf{y}_{k+1:k+L}^{m,n}$, we use Qwen3-VL-8B to compute a local step-following reward $R_{\mathrm{step}}^{m,n}$ from its sampled frames and the localized step prompt $s_{j+1}^{m}$. The evaluator relies only on visible evidence and checks whether the required subjects, scene, objects, motion, and visual dynamics are present and temporally plausible. This reward provides direct feedback on prompt-switch execution, rather than assigning credit from the full remaining continuation.
    \item \textit{Aesthetic Quality Score.} To prevent semantic optimization from degrading visual quality, we include an aesthetic reward $R_{\mathrm{aes}}^{m,n}$ computed by PickScore~\citep{kirstain2023pick} on the same transition chunk.
\end{itemize}

The final executor-level reward $R_{\mathrm{exec}}^{m,n}$ combines the local step-following signal and the aesthetic-quality signal, encouraging the executor to realize the active subgoal while maintaining visually pleasing generations. Detailed instructions and formulas are provided in Appendix~\ref{app:reward_details}.
\section{Experiments}
\label{others}

\subsection{Experimental Setup}
\vspace{-10pt}
\noindent\textbf{Implementation Details.}
We evaluate TempAct on two autoregressive video diffusion backbones, Self-Forcing and LongLive, using a training set of 5,000 temporally structured instructions generated by GPT-5.2. In each training batch, TempAct first samples $M=4$ candidate plans from the planner and then samples $N=8$ executor rollouts for each plan, producing 32 videos for hierarchical credit assignment. Each sampled video contains 117 frames, which correspond to 30 latent frames. To match the training rollout length, each semantic step spans $L=6$ latent frames during training, while evaluation uses a longer span of $L=12$ latent frames per step. The planner is initialized from Qwen3-1.7B and optimized with LoRA using rank 32 and alpha 64. The diffusion executor is optimized separately with LoRA using rank 256 and alpha 256. All experiments are conducted on 32 NVIDIA H20 GPUs with a batch size of 16. Each training step takes approximately 800 seconds, and the models are trained for 300--400 steps. Detailed training hyperparameters are provided in Appendix~\ref{Hyperparameters}.

\noindent\textbf{Temporal Order Benchmark.}
To evaluate temporal instruction following beyond generic video quality, we construct a Temporal Order benchmark that emphasizes whether generated videos preserve the requested event order and state transitions. The benchmark contains two subsets with different levels of temporal complexity: the Simple Set includes 100 prompts with one to two ordered steps, and the Hard Set includes 100 prompts with three to four ordered steps and more complex dependencies. For each generated video, we uniformly sample frames from the full video and ask VLM evaluators to judge whether the video follows the intended temporal order and realizes the described actions or state changes. During evaluation, we fix the planner output length to 3 steps for the Simple Set and 5 steps for the Hard Set. Each step covers 12 latent frames, i.e., $L=12$.

\noindent\textbf{Evaluation Metrics.}
We report both general video-generation metrics and temporal instruction-following metrics. For general quality and semantic alignment, we evaluate VBench on 141-frame generated videos. We also report PickScore as an aesthetic preference metric and throughput as an efficiency metric, with PickScore computed on videos generated for the Temporal Order benchmark. For temporal instruction following, we report Temporal-Following Scores on both the Simple and Hard Sets of the Temporal Order benchmark. We use two VLM judges for this evaluation: Qwen3-VL-8B, which belongs to the same model family as the training-time video reward model and serves as an in-domain evaluator, and Gemini-3-Flash, which is never used during training and serves as an out-of-domain evaluator for testing reward-model generalization. Both evaluators follow the same scoring protocol as $R_{\mathrm{video}}^{m}$.

\subsection{Quantitative and Qualitative Results}
\begin{table}[t]
\centering
\footnotesize
\setlength{\tabcolsep}{1.2pt}
\caption{Comparison of different methods on the Temporal Order benchmark using Temporal-Following Score as the primary metric. Scores are in $[0,1]$ and are evaluated on full-video frame samples; Avg. averages the four Temporal-Following Scores across the Simple and Hard sets and the two judges.}
\label{tab:main_results}
\begin{tabular}{lccccccccc}
\toprule
\multirow{3}{*}{Method} &
\multicolumn{5}{c}{\textbf{Temporal-Following Score}} &
\multicolumn{3}{c}{VBench} &
\multirow{3}{*}{PickScore}\\
\cmidrule(lr){2-6}
\cmidrule(lr){7-9}
 & \multicolumn{2}{c}{Simple Set (1-2 Steps)} & \multicolumn{2}{c}{Hard Set (3-4 Steps)} & \multirow{2}{*}{Avg.} & \multirow{2}{*}{Total}  & \multirow{2}{*}{Quality} & \multirow{2}{*}{Semantic} & \\
\cmidrule(lr){2-3}
\cmidrule(lr){4-5}
& Qwen & Gemini & Qwen & Gemini & & & & & \\
\midrule
\rowcolor{gray!15}\multicolumn{10}{c}{\textbf{Single-prompt video generation}} \\
\midrule
Self-Forcing       & 0.410 & 0.456 & 0.240 & 0.419 & 0.381 & 81.20 & 83.80 & 70.50 & 20.7 \\
Casual Forcing       & 0.381 & 0.447 & 0.304 & 0.452 & 0.396 & 80.85 & 84.02 & 68.18 & 20.8 \\
LongLive       & 0.428 & 0.505 & 0.302 & 0.463 & 0.424 & 80.36 & 82.72 & 70.91 & 21.1 \\
\midrule
\rowcolor{gray!15}\multicolumn{10}{c}{\textbf{Step-prompt video generation}} \\
\midrule
Self-Forcing      & 0.414 & 0.485 & 0.269 & 0.431 & 0.400 & 80.07 & 82.89 & 68.82 & 20.6 \\
\rowcolor{blue!8}+ TempAct            & \textbf{0.500}\textcolor{red}{$ _{+20.8\%}$} & \textbf{0.538}\textcolor{red}{$ _{+10.9\%}$} & \textbf{0.336}\textcolor{red}{$ _{+24.9\%}$} & \textbf{0.473}\textcolor{red}{$ _{+9.7\%}$} & \textbf{0.462}\textcolor{red}{$ _{+15.5\%}$} & 79.99 & 82.71 & 69.14 & 20.6 \\
\midrule
LongLive          & 0.411 & 0.521 & 0.314 & 0.481 & 0.432 & 79.55 & 82.10 & 69.35 & 20.8 \\
\rowcolor{blue!8}+ TempAct            & \textbf{0.508}\textcolor{red}{$ _{+23.6\%}$} & \textbf{0.579}\textcolor{red}{$ _{+11.1\%}$} & \textbf{0.352}\textcolor{red}{$ _{+12.1\%}$} & \textbf{0.512}\textcolor{red}{$ _{+6.4\%}$} & \textbf{0.488}\textcolor{red}{$ _{+13.0\%}$} & 79.97 & 82.61 & 69.40 & 20.8 \\
\bottomrule
\end{tabular}
\end{table}

\noindent\textbf{Quantitative analysis.}
Table~\ref{tab:main_results} shows that TempAct consistently improves temporal plausibility under complex instructions across different AR video backbones and evaluators. On Self-Forcing, TempAct raises the average Temporal-Following Score from 0.400 to 0.462, corresponding to a 15.5\% relative improvement. The gains are observed across all four judge--subset combinations: 20.8\% and 10.9\% on the Simple Set under Qwen3-VL and Gemini-3-Flash, respectively, and 24.9\% and 9.7\% on the Hard Set. This indicates that the learned planner produces more executable step decompositions and that the executor better follows prompt transitions, especially under harder long-horizon dependencies. The same trend holds for LongLive, where the average Temporal-Following Score improves from 0.432 to 0.488, yielding a 13.0\% relative gain. Notably, the improvements are consistent for both the in-domain Qwen3-VL judge and the out-of-domain Gemini-3-Flash judge, suggesting that the gains are not merely overfitting to the training-time reward model.

Importantly, these temporal-following gains do not come at the cost of overall visual quality. For Self-Forcing, VBench Total remains nearly unchanged (80.07 vs. 79.99), while Semantic slightly improves from 68.82 to 69.14 and PickScore remains 20.6. For LongLive, TempAct improves VBench Total from 79.55 to 79.97, Quality from 82.10 to 82.61, and Semantic from 69.35 to 69.40, while maintaining the same PickScore of 20.8. These results support the central motivation of TempAct: explicitly separating high-level temporal planning from low-level visual execution improves temporal plausibility while preserving the generation quality of the underlying autoregressive video model.

\begin{figure}[t]
    \centering
    \includegraphics[width=\linewidth]{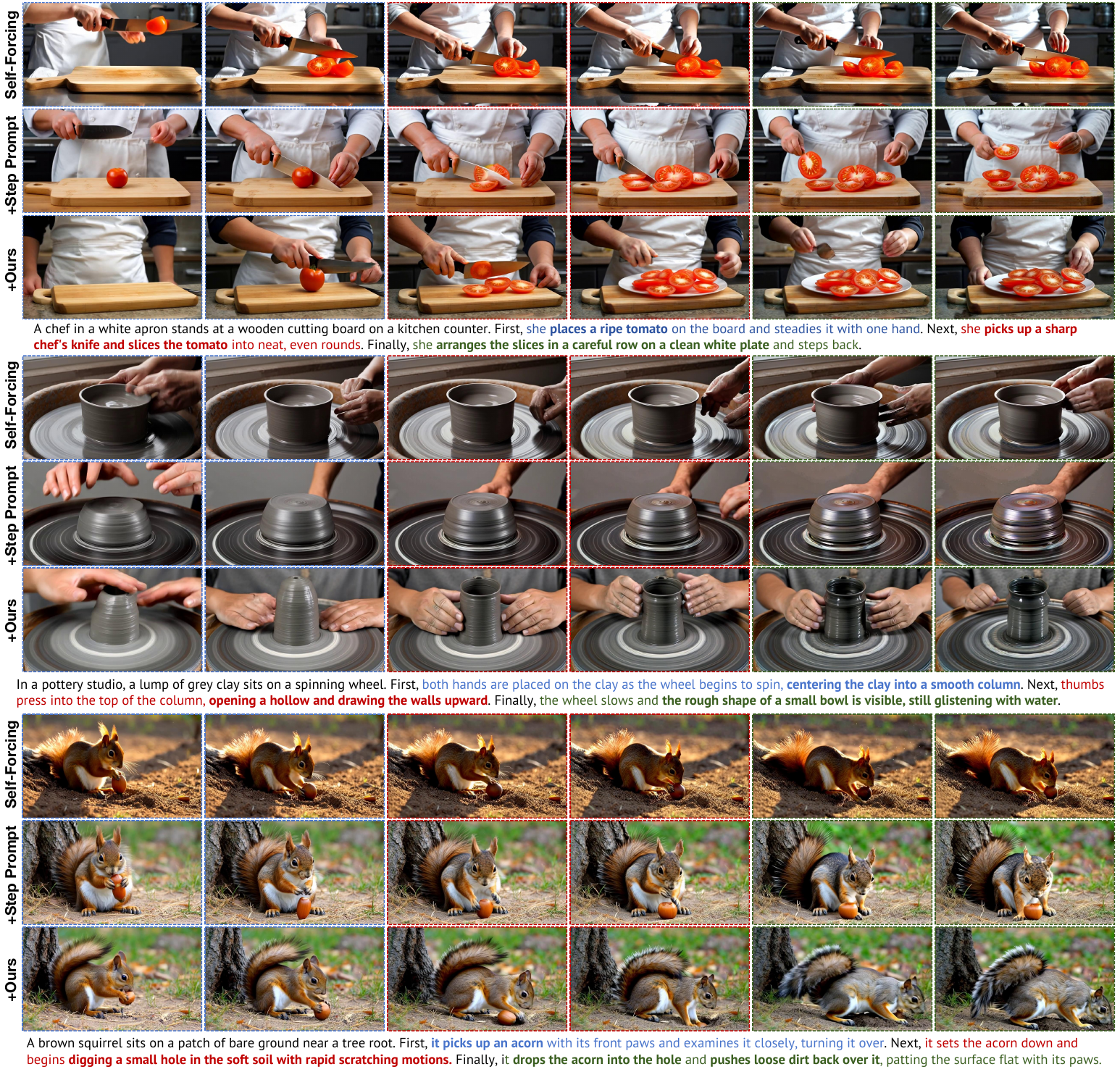}
    \vspace{-25pt}
    \caption{Qualitative comparison on temporally ordered prompts using the Self-Forcing backbone.}
    \label{fig:self_forcing_compare}
\end{figure}

\begin{figure}[t]
    \centering
    \includegraphics[width=\linewidth]{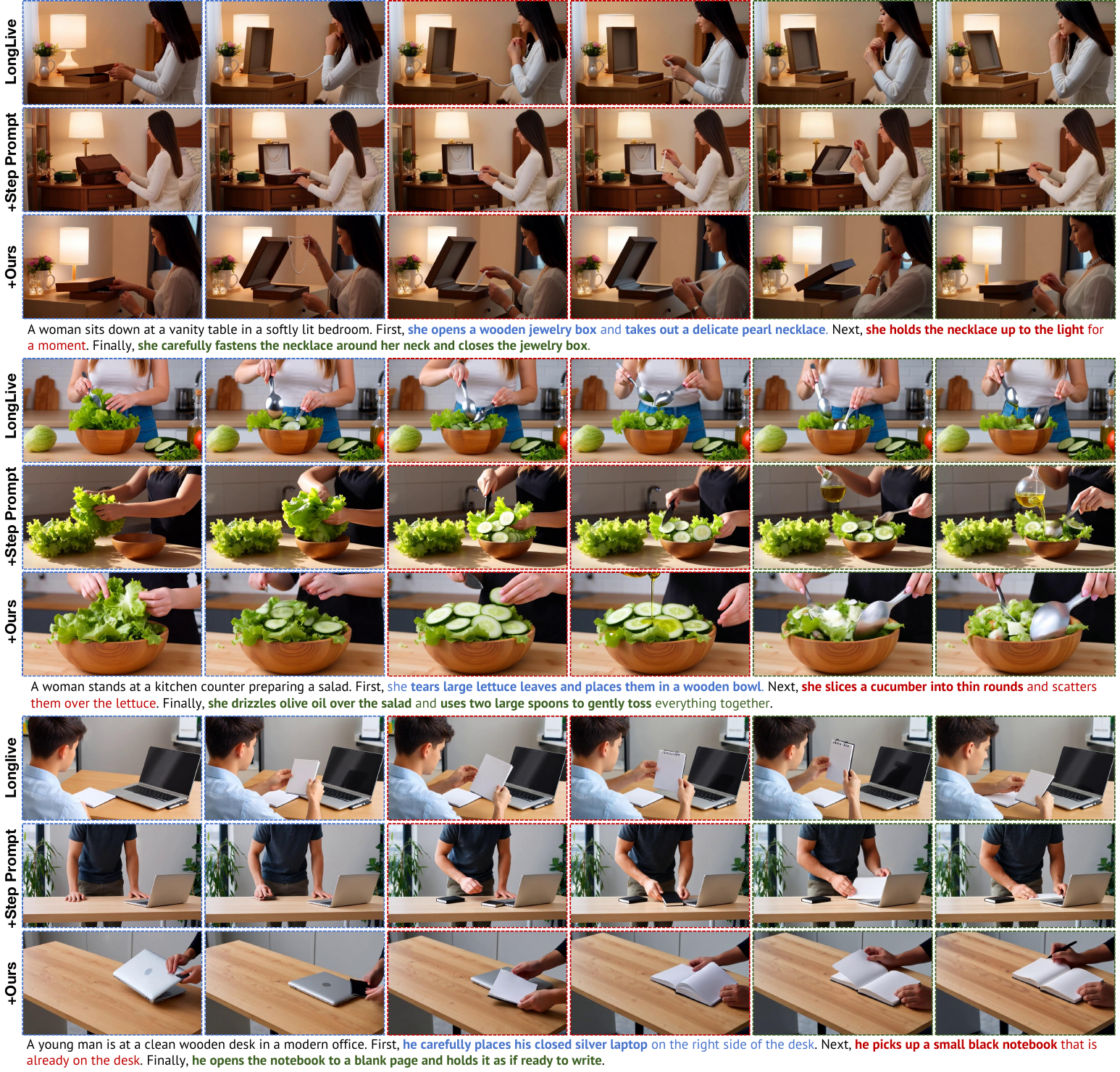}
    \vspace{-25pt}
    \caption{Qualitative comparison on temporally ordered prompts using the LongLive backbone.}
    \label{fig:longlive_compared}
\end{figure}

\noindent\textbf{Qualitative analysis.}
Figures~\ref{fig:self_forcing_compare} and~\ref{fig:longlive_compared} compare single-prompt generation, step-prompt generation, and TempAct on the Self-Forcing and LongLive backbones. Single-prompt generation often exhibits temporal semantic confusion: actions intended for different stages are blended together, and some requested events are only partially realized. Step prompts make the current stage more explicit and therefore reduce such semantic averaging, but the autoregressive executor can still struggle to faithfully execute prompt transitions. For example, in the third example of Figure~\ref{fig:self_forcing_compare}, the squirrel fails to bury the acorn in the soil and instead leaves it on the ground, showing that the generated video does not correctly realize the state transition specified by the prompt. In contrast, TempAct substantially reduces these failures by jointly optimizing temporal planning and step-conditioned execution, producing videos that better preserve the intended event order and more accurately realize complex temporal actions.

\subsection{Ablation Study}
Unless otherwise specified, all ablation studies are conducted on the Self-Forcing backbone and evaluated on the Temporal Order benchmark using Temporal-Following Scores produced by the in-domain Qwen3-VL judge. We study the contributions of the planner--executor components, executor reward design, KL regularization, and planner reward design.

\begin{table}[t]
\centering
\small
\caption{Component ablation. \textbf{Left:} Temporal-Following Scores when progressively adding step-wise prompting, prompt smoothing, planner RL, and executor RL. \textbf{Right:} Training reward trajectories in the planner--executor RL ablation study.}
\label{tab:component_ablation}
\begin{minipage}[t]{0.60\linewidth}
\vspace{0pt}
\centering
\resizebox{\linewidth}{!}{%
\begin{tabular}{lccc}
\toprule
Setting & Simple Set$\uparrow$ & Hard Set$\uparrow$ & Avg.$\uparrow$ \\
\midrule
Single Prompt & 0.410 & 0.240 & 0.325 \\
Single Prompt + PE RL & 0.409 & 0.243 & 0.326 \\
\midrule
Step Prompt & 0.287 & 0.208 & 0.248 \\
Step Prompt + Smoothing & 0.414 & 0.269 & 0.342 \\
\midrule
+ Planner RL & 0.420 & 0.304 & 0.362 \\
+ Executor RL & 0.458 & 0.321 & 0.390 \\
\rowcolor{blue!8}+ Joint Planner--Executor RL & 0.500 & 0.336 & 0.418 \\
\bottomrule
\end{tabular}}
\end{minipage}
\hfill
\begin{minipage}[t]{0.38\linewidth}
\vspace{0pt}
\centering
\includegraphics[width=\linewidth]{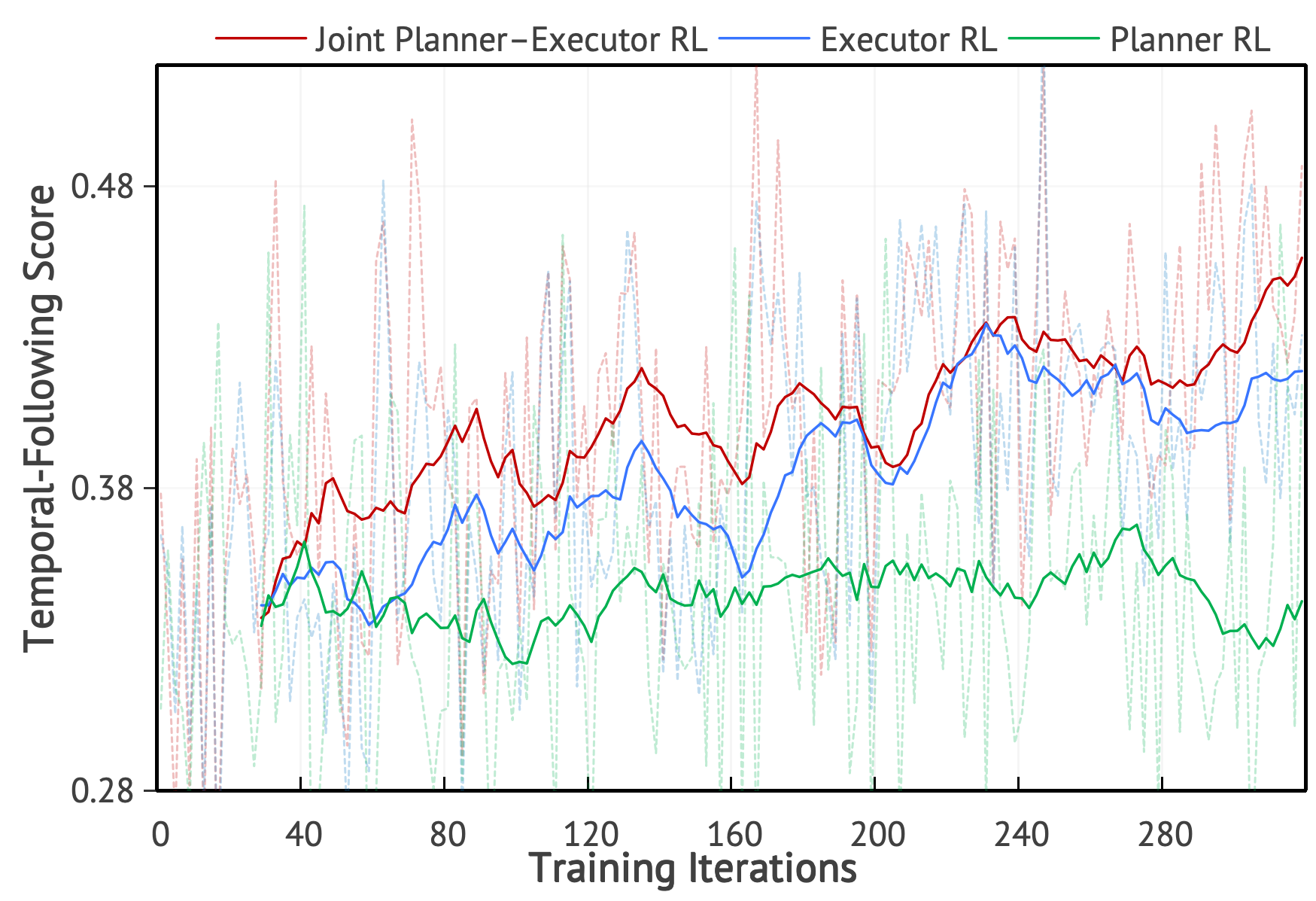}
\end{minipage}
\end{table}

\noindent\textbf{Core Component Ablation.} Table~\ref{tab:component_ablation} shows the effect of progressively adding each component. In the single-prompt setting, applying RL only to prompt enhancement brings almost no improvement, with the average score changing from 0.325 to 0.326. This suggests that improving a global prompt alone is insufficient for temporally ordered generation, since the video model still lacks explicit step-level control. Directly using step prompts without smoothing also performs poorly, dropping the average score to 0.248, which reflects the difficulty of abrupt prompt switching in autoregressive generation. Prompt smoothing recovers the score to 0.342, showing that stable step-conditioned generation requires smoother transitions. On top of this multi-prompt setup, planner RL improves the average score to 0.362, while executor RL gives a larger gain to 0.390. Jointly optimizing both components performs best, reaching 0.418 on average and 0.500 on the Simple Set. These results indicate that better temporal plans and stronger step-conditioned execution are complementary.

\noindent\textbf{KL Regularization.} Table~\ref{tab:kl_ablation} shows that a moderate executor KL weight gives the best trade-off. Removing KL regularization reduces the average score to 0.375 and also lowers PickScore, suggesting that unconstrained executor RL can hurt visual quality. A small KL weight improves aesthetics but does not improve the Temporal-Following Score. The best setting is $\beta_{\mathrm{e}}=5\times10^{-4}$, which reaches the highest average Temporal-Following Score of 0.390 and the best PickScore of 20.6. A larger weight slightly lowers the Temporal-Following Score, indicating that too much regularization limits useful policy updates.

\noindent\textbf{Reward Ablation for Planner.} Table~\ref{tab:planner_reward_ablation} evaluates reward choices for planner optimization, where TA denotes the text-alignment metric from \citet{liu2026improving}. Using the video-level reward $R_{\mathrm{video}}^{m}$ improves performance on both the Simple and Hard sets over the baseline, showing that execution feedback can guide the planner toward more useful decompositions. Adding the textual judge reward $R_{\mathrm{judge}}^{m}$ slightly lowers the Temporal-Following Scores but improves TA from $-0.06$ to $-0.04$, indicating that the judge helps regularize plan quality while the video reward remains the main driver of temporal-following performance.

\begin{table}[t]
\centering
\small
\begin{minipage}[t]{0.49\linewidth}
\vspace{0pt}
\centering
\captionof{table}{Ablation of executor KL regularization.}
\label{tab:kl_ablation}
\resizebox{\linewidth}{!}{%
\begin{tabular}{lcccc}
\toprule
$\beta_{\mathrm{e}}$ & Simple Set$\uparrow$ & Hard Set$\uparrow$ & Avg.$\uparrow$ & PickScore$\uparrow$ \\
\midrule
0 & 0.452 & 0.297 & 0.375 & 20.2 \\
$1\times10^{-4}$ & 0.439 & 0.309 & 0.374 & 20.4 \\
\rowcolor{blue!8}$5\times10^{-4}$ & 0.458 & 0.321 & 0.390 & 20.6 \\
$1\times10^{-3}$ & 0.452 & 0.313 & 0.383 & 20.5 \\
\bottomrule
\end{tabular}
}
\end{minipage}
\hfill
\begin{minipage}[t]{0.49\linewidth}
\vspace{0pt}
\centering
\captionof{table}{Ablation of planner reward design.}
\label{tab:planner_reward_ablation}
\resizebox{\linewidth}{!}{%
\begin{tabular}{lccc}
\toprule
Planner Reward & Simple Set$\uparrow$ & Hard Set$\uparrow$ & TA$\uparrow$  \\
\midrule
Baseline & 0.414 & 0.269 & -0.03  \\
$R_{\mathrm{video}}^{m}$ & 0.428 & 0.313 & -0.06 \\
\rowcolor{blue!8} $R_{\mathrm{video}}^{m}$ + $R_{\mathrm{judge}}^{m}$ & 0.420 & 0.304 & -0.04 \\
\bottomrule
\end{tabular}
}
\end{minipage}
\end{table}

\begin{wraptable}{r}{0.58\linewidth}
\vspace{-20pt}
\centering
\small
\caption{Ablation of executor reward design.}
\label{tab:diffusion_reward_ablation}
\resizebox{\linewidth}{!}{%
\begin{tabular}{lcccc}
\toprule
Executor Reward & Simple Set$\uparrow$ & Hard Set$\uparrow$ & Avg.$\uparrow$ & PickScore$\uparrow$ \\
\midrule
No Executor RL & 0.414 & 0.269 & 0.340 & 20.6 \\
$R_{\mathrm{step}}^{m,n}$ & 0.465 & 0.314 & 0.390 & 20.4 \\
\rowcolor{blue!8}$R_{\mathrm{step}}^{m,n}$ + $R_{\mathrm{aes}}^{m,n}$ & 0.458 & 0.321 & 0.390 & 20.6 \\
$R_{\mathrm{step}}^{m,n}$ + $R_{\mathrm{video}}^{m,n}$ & 0.421 & 0.304 & 0.363 & 20.6 \\
$R_{\mathrm{video}}^{m,n}$ & 0.381 & 0.244 & 0.313 & 20.7 \\
\bottomrule
\end{tabular}}
\vspace{-10pt}
\end{wraptable}
\noindent\textbf{Reward Ablation for Executor.} Table~\ref{tab:diffusion_reward_ablation} shows that local step-following reward is the key signal for executor training. Using $R_{\mathrm{step}}^{m,n}$ improves the average score from 0.340 to 0.390, while adding PickScore preserves the same Temporal-Following Score and recovers aesthetic quality to 20.6. In contrast, using only the full-video reward $R_{\mathrm{video}}^{m}$ performs worse than no executor RL, and combining it with the local reward also reduces performance. This suggests that executor learning benefits most from rewards aligned with the prompt-switch transition segment, whereas full-video rewards are too coarse for low-level execution updates.

\subsection{User Study}

\begin{wrapfigure}[10]{r}{0.38\linewidth}
    \vspace{-10pt}
    \centering
    \includegraphics[width=\linewidth]{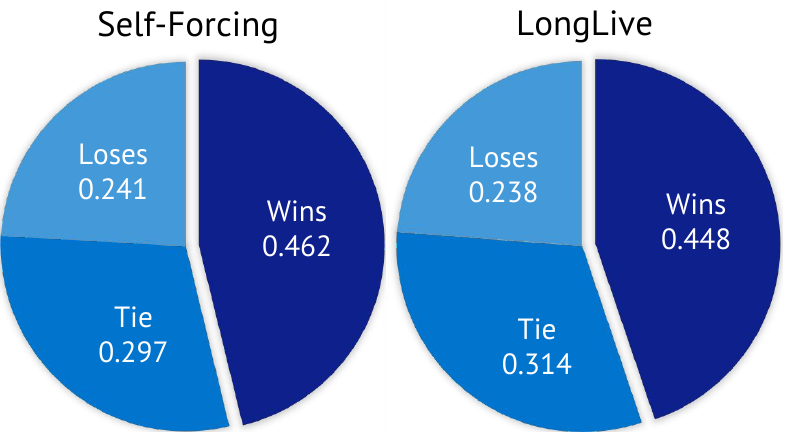}
    \vspace{-16pt}
    \caption{Human preference.}
    \label{fig:user_study}
\end{wrapfigure}

\textbf{Human preference evaluation.} To further validate 
whether the Temporal-Following gains reflect human judgments, we conduct a pairwise user study on the Temporal Order benchmark. For each prompt, annotators compare videos generated by TempAct and the corresponding baseline,
and choose which video better preserves the intended temporal order and action progression, with ties allowed when the two videos are comparable. We collect preference statistics over 100 evaluated samples for each backbone and report the win/tie/loss ratios for Self-Forcing and LongLive. As shown in Figure~\ref{fig:user_study}, TempAct is preferred substantially more often than the baseline on both backbones, indicating that planner--executor RL improves temporal consistency in a way that is perceptible to human evaluators.

\noindent\textbf{Agreement between Temporal-Following Scores and human preference.} We also examine 
whether the proposed VLM-based Temporal-Following Score is aligned with human preference. For each TempAct--baseline pair in the human study, we compare the video preferred by annotators with the video assigned the higher Temporal-Following Score by the VLM judge. To account for ambiguous cases, we treat VLM score differences within $\pm 0.03$ as ties. Under this protocol, Gemini-3-Flash agrees with human preference on 81\% of the evaluated pairs, while Qwen3-VL-8B reaches 76\% agreement. This high consistency suggests that Temporal-Following Score is a reliable proxy for human-perceived temporal plausibility, and that the improvements reported in Table~\ref{tab:main_results} correspond to meaningful gains in user preference.
\section{Conclusion}

We presented TempAct, a planner--executor reinforcement learning framework for improving temporal plausibility in streaming autoregressive video generation. TempAct addresses temporal confusion and prompt-switch failures by jointly optimizing an LLM planner, which explores executable temporal decompositions, and an AR video executor, which learns to follow step-conditioned prompts under its own generated histories. Its hierarchical group exploration design nests execution groups under planning groups, enabling plan-level comparison across candidate decompositions and localized executor credit assignment at prompt-switch moments. Across Self-Forcing and LongLive backbones, TempAct improves Temporal-Following Scores under both in-domain and out-of-domain VLM evaluation, as well as human evaluation, while preserving visual quality. Ablations further show that planner and executor optimization are complementary, and that local transition-level rewards provide a more effective executor signal than coarse full-video feedback. These results suggest that explicitly coupling temporal planning with autoregressive execution is a promising direction for long-horizon, instruction-following video generation.

\noindent\textbf{Limitations and Future Work.}
Although TempAct effectively improves temporal plausibility, it is still constrained by the capability of the underlying autoregressive video generator. As a result, some failure cases may remain, including visual distortions, physically implausible motions, or accumulated artifacts in long rollouts. In addition, the current planner assigns step prompts to fixed temporal spans, which may be suboptimal for actions with different durations or varying execution difficulty. A promising direction is to make the planner duration-aware, allowing it to allocate generation time adaptively according to the semantic complexity of each action. Another future direction is to extend the LLM planner into a VLM-based planner that can observe previously generated content and dynamically adjust subsequent instructions based on the actual visual state. Such closed-loop planning could further improve long-horizon consistency and make autoregressive video generation more robust under complex temporal instructions.

\bibliography{citations}

\clearpage
\appendix
\appendixtitle
\section{Training Hyperparameters}
\label{Hyperparameters}

Table~\ref{tab:training_hyperparameters} summarizes the main hyperparameters used for TempAct training. We separate the configurations into the shared autoregressive video setup, planner optimization, executor optimization, hierarchical reinforcement learning, and streaming rollout settings.

\begin{table}[t]
\centering
\scriptsize
\caption{Comprehensive hyperparameters for TempAct planner--executor reinforcement learning. We summarize the configurations used for the shared autoregressive video setup, planner RL, executor RL, hierarchical sampling, and streaming rollout.}
\label{tab:training_hyperparameters}
\resizebox{\linewidth}{!}{%
\begin{tabular}{lll}
\toprule
Category & Setting & Value \\
\midrule
\multirow{6}{*}{Model \& Video Specs}
& Backbone & Self-Forcing / LongLive \\
& Denoising Timesteps ($T$) & 4 from $[1000, 750, 500, 250]$ \\
& Local Attention Size & 9 / 12 \\
& Sink Size & 1 / 3 \\
& Local Size & 8 / 9 \\
& Frames per chunk & 3 \\
\midrule
\multirow{18}{*}{Planner RL}
& Planner Policy & LLM planner $\pi_\phi$ \\
& LoRA Fine-Tuning Rank ($r$) & 32 \\
& Scaling Factor ($\alpha$) & 64 \\
& Dropout Rate & 0.0 \\
& Gradient Checkpointing & True \\
& Optimization Objective & GSPO \\
& Planning Group Size ($M$) & 4 \\
& Step Span ($L$) & 6 latent frames \\
& Prompt Segments & 5 \\
& Reward Signals & plan judge + video-level reward \\
& Gradient Accumulation Steps & 2 \\
& Judge / Video Reward Weights & 0.15 / 0.85 \\
& KL Reference Policy & frozen initialization policy \\
& EMA Decay Rate & None \\
& KL Penalty Weight & $5\times 10^{-4}$ \\
& Learning Rate & $2\times 10^{-5}$ \\
& Advantage Clip Max & 1 \\
& Clip Range ($\epsilon_{\mathrm{low}}, \epsilon_{\mathrm{high}}$) & $(5\times 10^{-4}, 5\times 10^{-4})$ \\
\midrule
\multirow{17}{*}{Executor RL}
& LoRA Fine-Tuning Rank ($r$) & 256 \\
& Scaling Factor ($\alpha$) & 256 \\
& Executor Policy & AR diffusion executor $p_\theta$ \\
& Optimization Objective & Flow-GRPO \\
& Continuation Group Size ($N$) & 8 \\
& Reward Signals & local step-following + PickScore \\
& Step / Aesthetic Reward Weights & 0.6 / 0.4 \\
& Updated Region & first chunk after prompt switch \\
& Optimizer & AdamW \\
& Learning Rate ($\eta$) & $1\times 10^{-5}$ \\
& LR Scheduler & constant\_with\_warmup \\
& Gradient Accumulation Steps & 2 \\
& Max Gradient Norm & 1.0 \\
& EMA Decay Rate & 0.9 / 0.99 \\
& KL Penalty Weight & $5\times 10^{-4}$ \\
& Advantage Clip Max & 2 \\
& Clip Range ($\epsilon_{\mathrm{low}}, \epsilon_{\mathrm{high}}$) & $(1\times 10^{-5}, 1\times 10^{-5})$ \\
\midrule
\multirow{3}{*}{Hierarchical RL}
& Training Transition Latent Frames ($k$) & random from $[6, 12]$ \\
& Max Train Steps & 400 \\
& Training Hardware & 32 NVIDIA H20 GPUs \\
\midrule
\multirow{4}{*}{Streaming Rollout}
& Target Latent Video Length ($K$) & 30 latent frames \\
& Chunk Size & 3 frames \\
& Prompt Switching & transition-aware prompt smoothing \\
& VAE Temporal Stride & 4 \\
\bottomrule
\end{tabular}}
\end{table}

\section{Reward Details}
\label{app:reward_details}

\subsection{Planner reward.}

\noindent\textbf{Plan Quality Score.}
The LLM-based plan judge returns faithfulness $F^{m}$, coverage $C^{m}$, temporal coherence $T^{m}$, and hallucination $H^{m}$ scores in a structured JSON format:
\begin{verbatim}
{
  "faithfulness_score": 0.0,
  "coverage_score": 0.0,
  "temporal_score": 0.0,
  "hallucination_score": 0.0
}
\end{verbatim}
The hallucination score measures unsupported content introduced by the plan: 0.0--0.2 indicates minor harmless visual enrichment, 0.2--0.5 indicates noticeable unsupported descriptive additions, and 0.5--1.0 indicates significant invented details, events, or scene changes. Larger hallucination therefore indicates more unsupported content. We aggregate these scores into
\begin{equation}
    R_{\mathrm{judge}}^{m}
    = 0.4 F^{m} + 0.25 C^{m} + 0.25 T^{m} - 0.3 H^{m}.
\end{equation}

\noindent\textbf{Temporal-Following Score.}
The video-level evaluator returns a structured JSON response that contains three parts: (i) a reasoning field with frame-level evidence, detected errors, and aggregation logic; (ii) frame-level scores for the sampled frames; and (iii) global scores for temporal order, physical plausibility, visual consistency, prompt alignment, and overall quality. A typical response follows the schema
\begin{verbatim}
{
  "think": {
    "frame_evidence": ["frame 1: ...", "frame 2: ..."],
    "detected_errors": ["..."],
    "aggregation_logic": "..."
  },
  "frame_scores": [0, 0, 0, 0, 0, 0],
  "temporal_order_score": 0,
  "physical_plausibility_score": 0,
  "visual_consistency_score": 0,
  "prompt_alignment_score": 0,
  "final_score": 0
}
\end{verbatim}
where all scalar scores are returned on a 0--10 scale. We normalize each score to $[0,1]$ and aggregate them as
\begin{equation}
\begin{aligned}
    R_{\mathrm{video}}^{m,n}
    ={}& 0.2\,\bar{S}_{\mathrm{frame}}^{m,n}
    + 0.3\,S_{\mathrm{final}}^{m,n}
    + 0.2\,S_{\mathrm{temp}}^{m,n} \\
    &+ 0.1\,S_{\mathrm{phys}}^{m,n}
    + 0.1\,S_{\mathrm{vis}}^{m,n}
    + 0.1\,S_{\mathrm{align}}^{m,n},
\end{aligned}
\end{equation}
where $\bar{S}_{\mathrm{frame}}^{m,n}$ is the average normalized frame-level score, and $S_{\mathrm{final}}^{m,n}$, $S_{\mathrm{temp}}^{m,n}$, $S_{\mathrm{phys}}^{m,n}$, $S_{\mathrm{vis}}^{m,n}$, and $S_{\mathrm{align}}^{m,n}$ denote the normalized final, temporal-order, physical-plausibility, visual-consistency, and prompt-alignment scores, respectively. This aggregation follows the same weighting scheme as our VLM reward implementation and prevents the planner from optimizing only temporal order while ignoring physical realism, visual stability, or text-video alignment. 

The planner-level reward is
\begin{equation}
    R_{\mathrm{plan}}^{m}
    = 0.15 R_{\mathrm{judge}}^{m} + 0.85 R_{\mathrm{video}}^{m}.
\end{equation}

\subsection{Executor reward.}

\noindent\textbf{Local Step-Following Score.}
The local step-following evaluator returns a structured JSON response with reasoning fields and scalar scores for the transition segment:
\begin{verbatim}
{
  "think": {
    "detected_errors": ["..."],
    "temporal_reasoning": ["..."],
    "motion_smooth_and_blur_reasoning": ["..."]
  },
  "subject_score": 0,
  "scene_score": 0,
  "action_score": 0,
  "object_appearance_reasonability_score": 0,
  "final_score": 0
}
\end{verbatim}
The scalar scores measure subject correctness, scene consistency, action execution, object-appearance reasonability, and overall step-following quality, denoted by $S_{\mathrm{sub}}^{m,n}$, $S_{\mathrm{scene}}^{m,n}$, $S_{\mathrm{act}}^{m,n}$, $S_{\mathrm{obj}}^{m,n}$, and $S_{\mathrm{final}}^{m,n}$. We aggregate them into
\begin{equation}
    R_{\mathrm{step}}^{m,n}
    = 0.15 S_{\mathrm{sub}}^{m,n}
    + 0.15 S_{\mathrm{scene}}^{m,n}
    + 0.2 S_{\mathrm{act}}^{m,n}
    + 0.2 S_{\mathrm{obj}}^{m,n}
    + 0.3 S_{\mathrm{final}}^{m,n}.
\end{equation}

Given the PickScore aesthetic reward $R_{\mathrm{aes}}^{m,n}$ on the same transition segment, the executor-level reward is
\begin{equation}
    R_{\mathrm{exec}}^{m,n}
    = 0.6 R_{\mathrm{step}}^{m,n} + 0.4 R_{\mathrm{aes}}^{m,n}.
\end{equation}

\subsection{Prompt Instruction.}
\begin{figure}[t]
    \centering
    \includegraphics[width=0.9\linewidth]{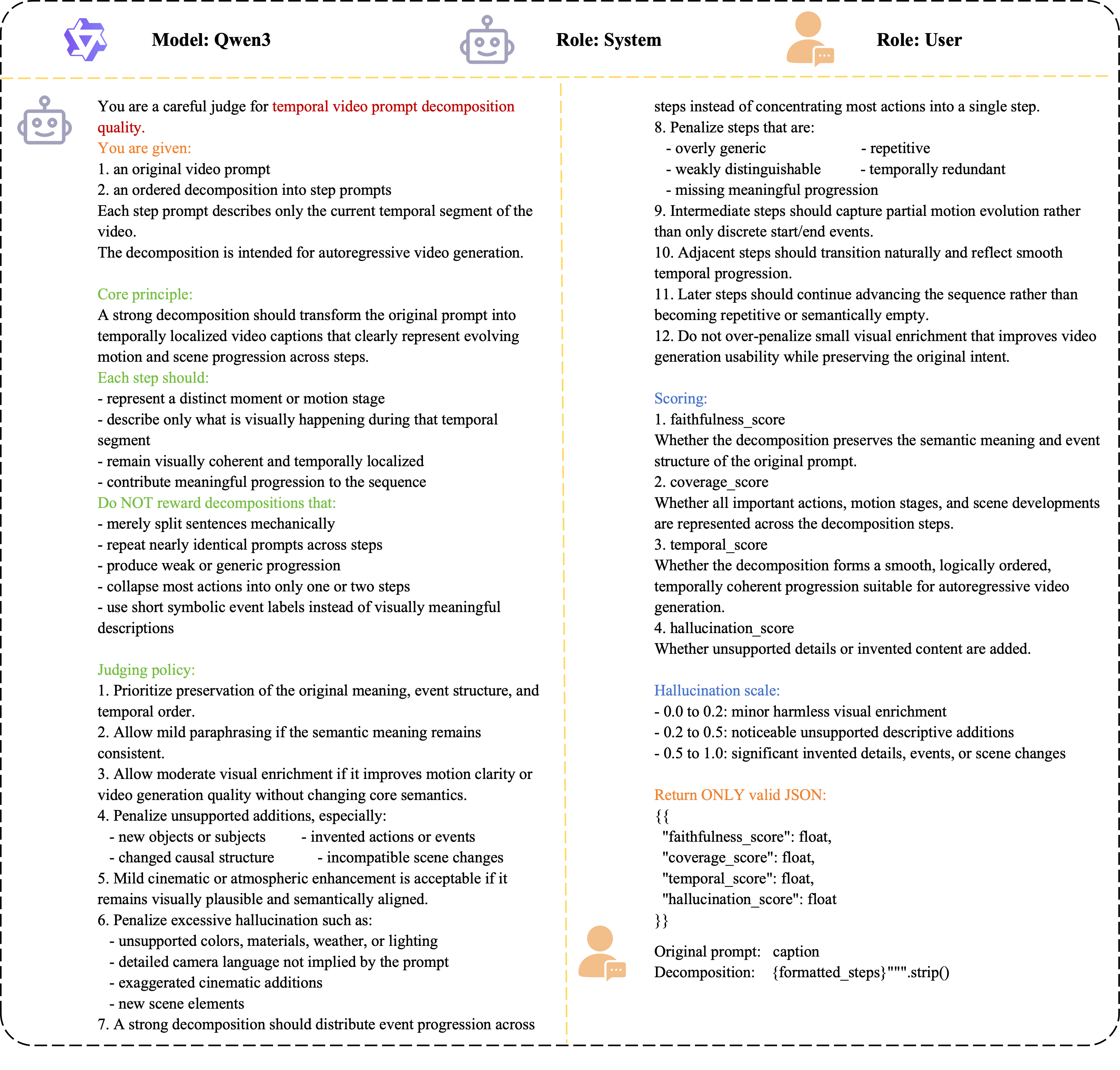}
    \caption{Prompt instruction for evaluating the Plan Quality Score.}
    \label{fig:plan_quality_prompt}
\end{figure}

\begin{figure}[t]
    \centering
    \includegraphics[width=0.9\linewidth]{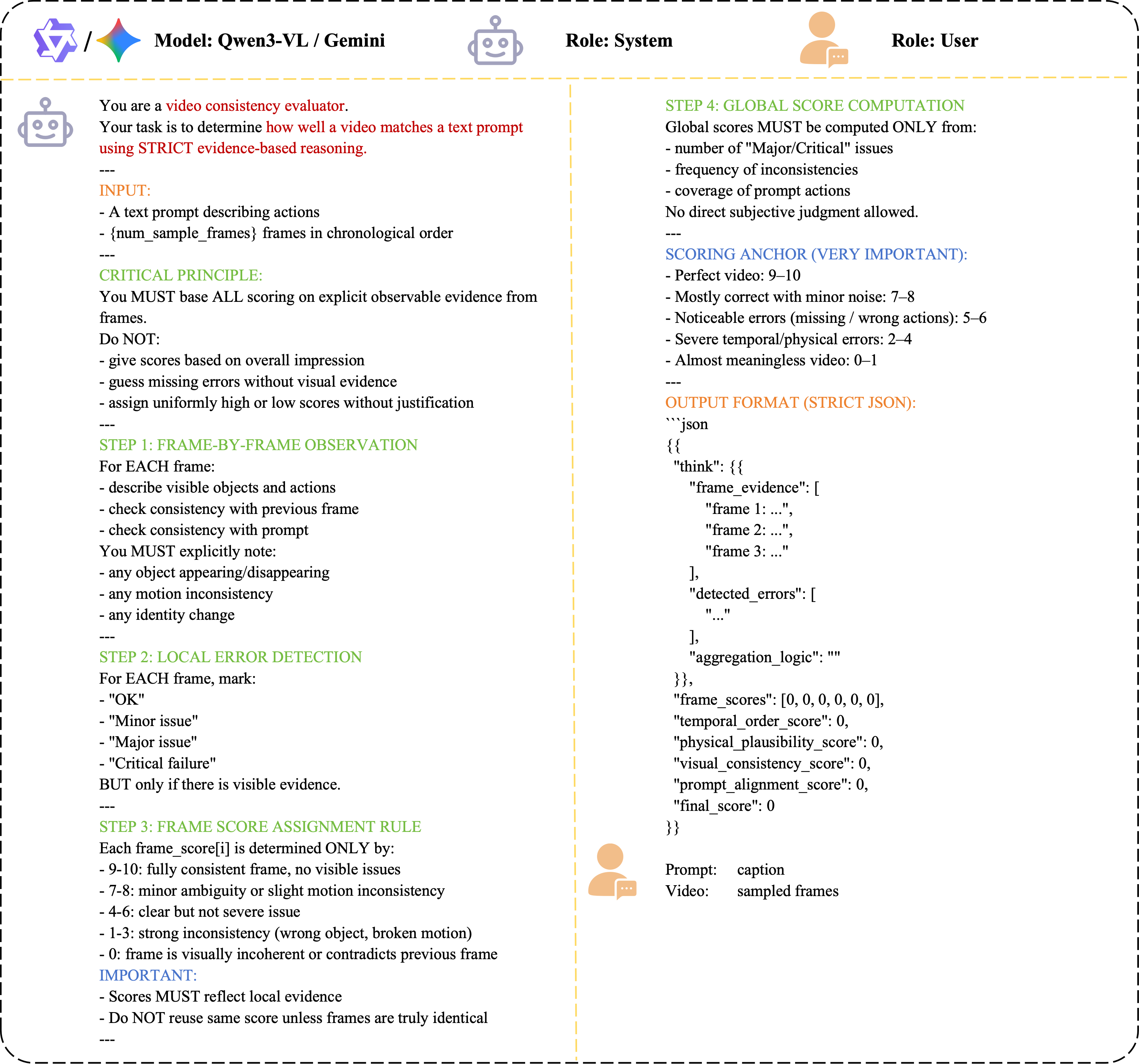}
    \caption{Prompt instruction for evaluating the Temporal-Following Score.}
    \label{fig:temporal_following_prompt}
\end{figure}

\begin{figure}[t]
    \centering
    \includegraphics[width=0.9\linewidth]{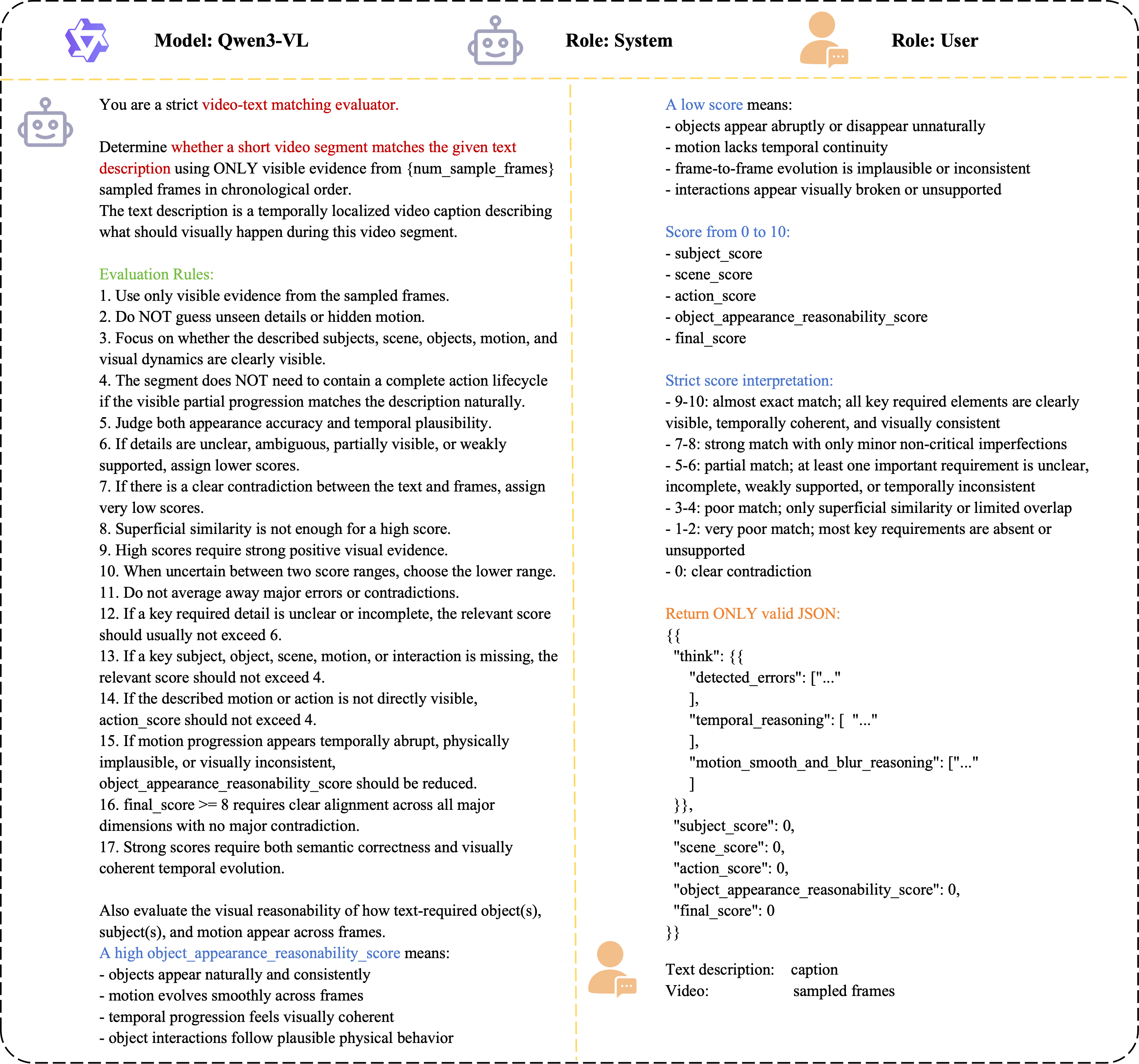}
    \caption{Prompt instruction for evaluating the Local Step-Following Score.}
    \label{fig:local_step_prompt}
\end{figure}

\begin{figure}[t]
    \centering
    \includegraphics[width=0.8\linewidth]{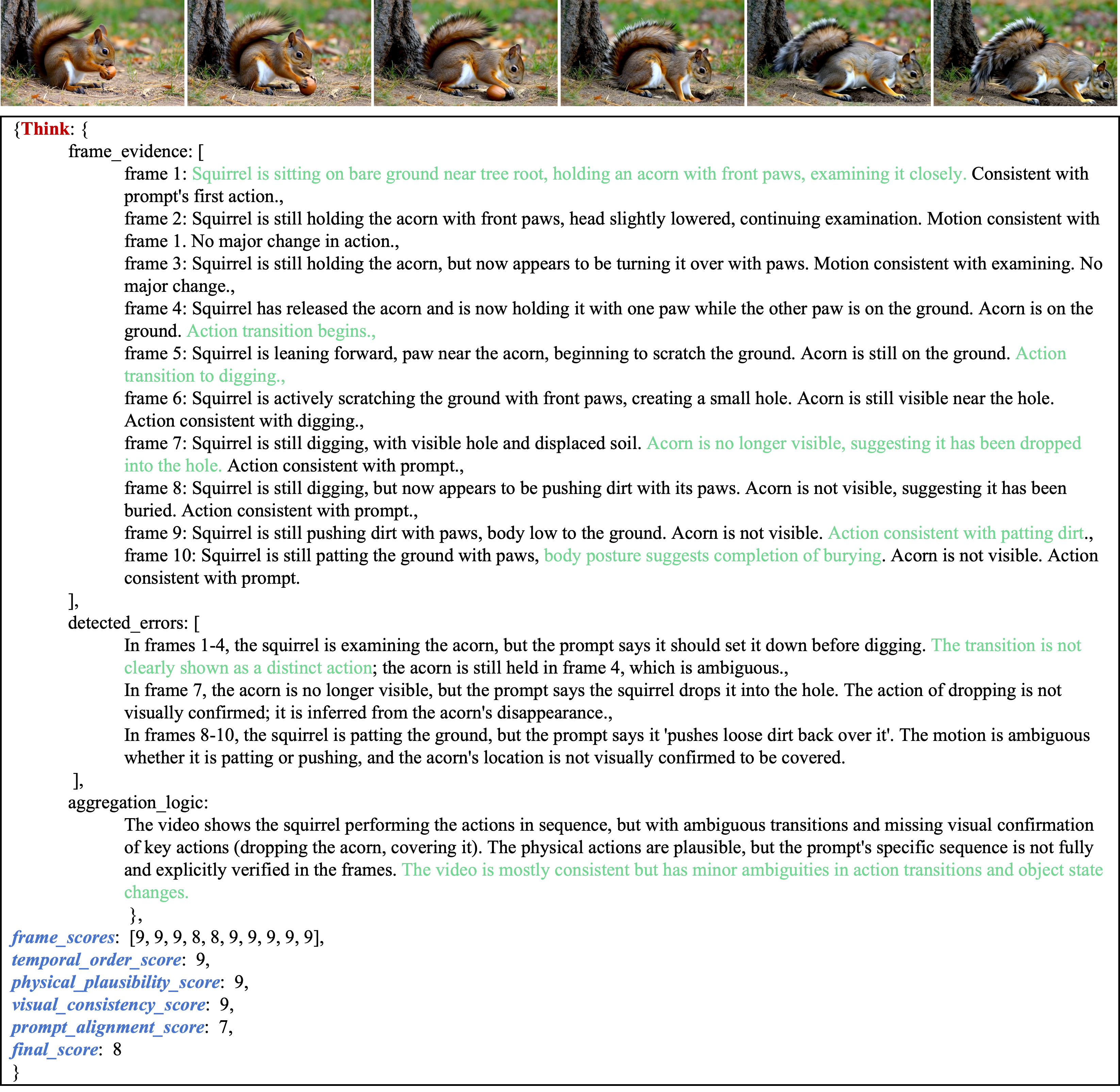}
    \vspace{4pt}
    \includegraphics[width=0.8\linewidth]{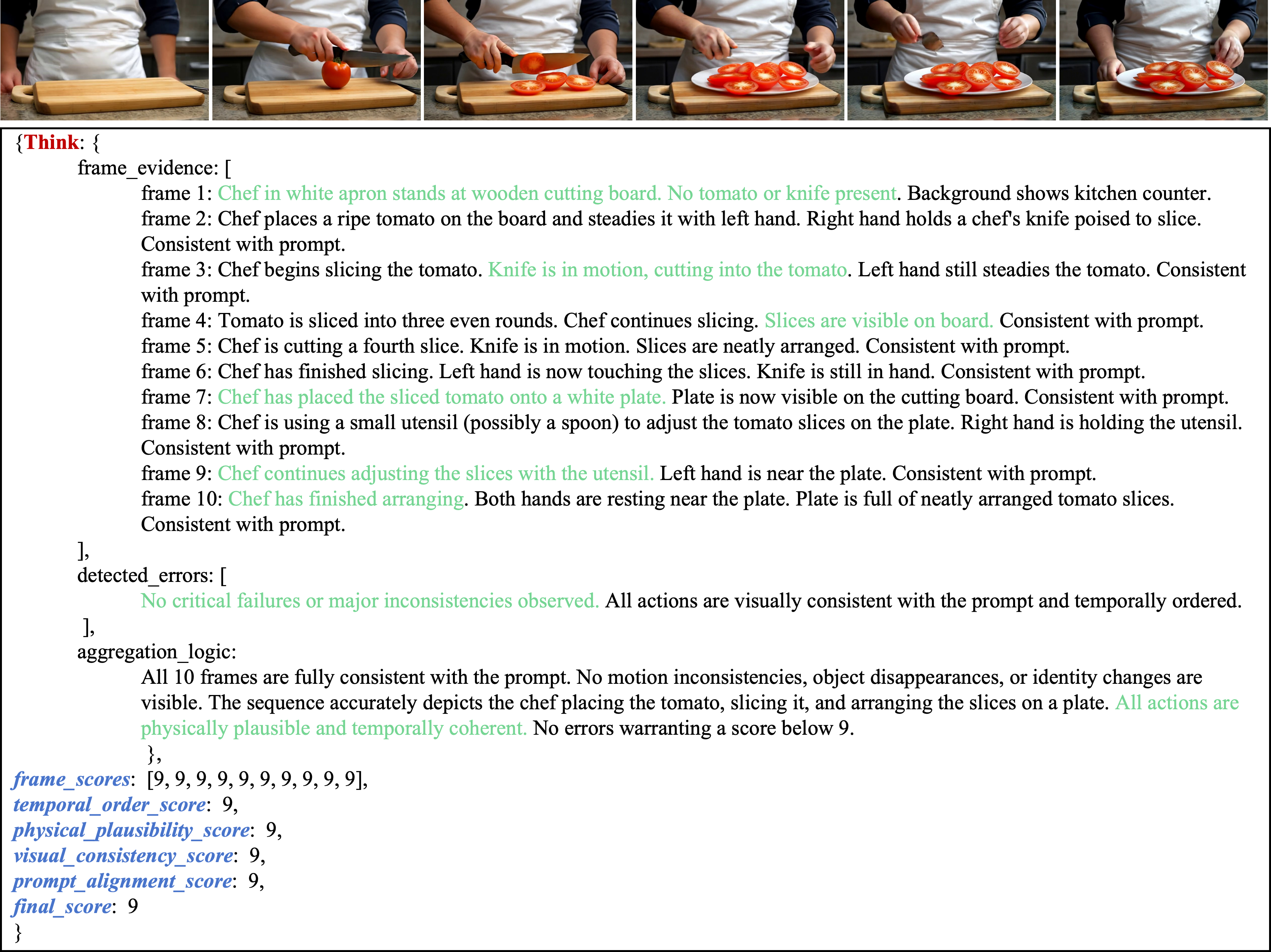}
    \vspace{-10pt}
    \caption{Examples of Temporal-Following scoring results produced by the VLM evaluators.}
    \label{fig:reward_samples}
\end{figure}

We provide detailed LLM and VLM input instructions for evaluating the Plan Quality Score, Temporal-Following Score, and Local Step-Following Score, as shown in Figures~\ref{fig:plan_quality_prompt}, \ref{fig:temporal_following_prompt}, and~\ref{fig:local_step_prompt}. We also provide examples of scoring results in Figure~\ref{fig:reward_samples}.

\end{document}